\documentclass[a4paper, 11pt]{article}
\usepackage{framed}
\usepackage{multicol}
\usepackage{float}
\usepackage{epsfig,fancyhdr}
\usepackage{amsmath,amssymb}
\usepackage{bbm}
\usepackage{breqn}
\usepackage{scrextend}
\usepackage[british]{babel}
\usepackage{dirtytalk}
\usepackage{subfig}
\usepackage{colortbl}
\usepackage{url,graphicx,tabularx,array,geometry}
\usepackage{epstopdf}
\usepackage{mathrsfs}
\usepackage{hyperref}
\usepackage{cite}
\usepackage[utf8]{inputenc}
\usepackage[]{algpseudocode}
\usepackage[]{algorithm}
\usepackage[toc]{appendix}
\newtheorem{theorem}{Theorem}[section]

\usepackage{amssymb}
\DeclareMathAlphabet{\mathpzc}{OT1}{pzc}{m}{it}

\DeclareMathOperator*{\argmin}{arg\,min}
\newcommand{\Var}{\mathrm{Var}}

\definecolor{light-gray}{gray}{0.95}

\definecolor{beige}{rgb}{0.96, 0.96, 0.86}

\begin{document}
\title{\textbf{On Recovering Latent Factors From Sampling And Firing Graph}}

\author{
{\color{white} a}\\
Pierre Gouedard\\
{\color{white} a}\\
}
\maketitle

\setlength{\parindent}{0pt}

\begin{abstract}
Consider a set of latent factors whose observable effect of activation is caught on a measure space that appears as a grid of bits tacking value in $\{0, 1 \}$. This paper intend to deliver a theoretical and practical answer to the question: Given that we have access to a perfect indicator of the activation of latent factors that label a finite dataset of grid's activity, can we imagine a procedure to build a generic identificator of factor's activations ?
\end{abstract}

\section{Introduction}

This paper starts by introducing a mathematical framework for our solution. Then it describes a procedure to build the generic factor's activations identificator. Finally it presents the result of the procedure for two particular statistical modelling of the measure grid's activity. This paper has been influenced by modern machine learning techniques, reviewed in \cite{PCML-1}, especially algorithm that perform automated feature engineering such as neural network and deep learning  \cite{PCML-2} as well as tree learning techniques \cite{PCML-3} and improvements \cite{PCML-4} and \cite{PCML-5}. Finally, modern signal processing techniques, that I have been taught at the Ecole Polytechnique F\'ed\'erale De Lausanne, reviewed in \cite{SP-1}, and recent work in statistics in large dimensions, to which I have been introduced during my stay in the Laboratoire d'Informatique Gaspard Monge, reviewed in \cite{TAO-1}, has been more than determinant for the conception of this paper. In order to make the core subject of this report more consistent, we introduce the following notations:

 \begin{figure}[h]
 \centering
   \includegraphics[scale=0.25]{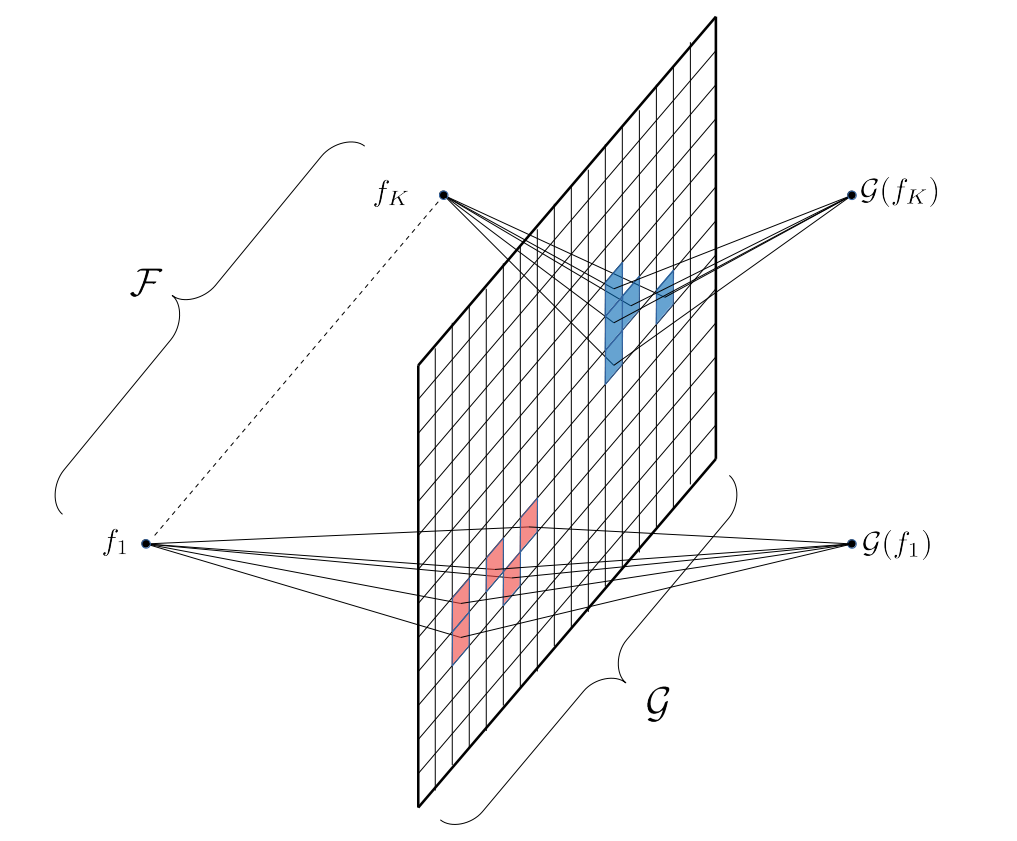}
 \caption{Measure grid model}
 \label{fig:model_overview}
\end{figure}

\begin{itemize}
\item $n$ the size of the measure grid. 
\item $\mathcal{G}$ the measure grid, composed of bits, $\mathcal{G} = \lbrace b_1, \ldots b_n \rbrace$.
\item $S(\mathcal{G})$  the set of all possible permutations of $\mathcal{G}$, $\vert S(\mathcal{G})\vert = 2^{n}$.
\item $S(\mathcal{G}, l)$  the set of all permutations of $\mathcal{G}$ of size $l$.
\item $K$ the number of latent factors.
\item $\mathcal{F}$ the set of latent factors, $\mathcal{F} = \lbrace f_1, \ldots f_K \rbrace$.
\item $S(\mathcal{F})$  the set of all possible permutations of $\mathcal{F}$, $\vert S(\mathcal{F})\vert = 2^{K}$.
\item $S(\mathcal{F}, l)$  the set of all permutations of $\mathcal{F}$ of size $l$.
\item $\mathcal{G}(f)$  the set of bits activated by factor $f$, $\mathcal{G}(f) \in S(\mathcal{G})$ and $f \in \mathcal{F}$.  
\item $\mathcal{G}^{-1}(b)$  the set of factors that activate grid's bit $b$, $\mathcal{G}^{-1}(b) \in S(\mathcal{F})$ and $b \in \mathcal{G}$.
\item $F(2)$ the field with elements $\{0, 1\}$, equipped with the logical XOR and logical AND respectively as the addition and multiplication.
\end{itemize}

\section{Definitions And Properties}

\subsection{Statistical definitions}

In this section we provide formalism for the statistical description of factors's activity and their signature on the measure grid.

\subsubsection*{Activation of factors}

Each factor takes value in $\{ 0, 1\}$ at each instant of time. A factor with value 1 at some instant is active, otherwise it has value 0. At this stage of the paper we assume no particular statistical model for factors. Nevertheless, if we consider the set of all possible combination of active and unactive factors ($F(2)^{K}$), we assume that there is a well defined distribution $d'$ such that 

\begin{align*}
d' &: F(2)^{K} \rightarrow [0,1]\\
d'_x &= \mathbb{P}\left(f_1 = x_1, \ldots, f_K = x_K \right)
\end{align*}

The statistical signature of a factor on the measure grid describes how the factor is linked to measure grid's bits. At this stage we simply assume that there is a well defined probability measure so that for any $I \in S(\mathcal{G})$

\begin{align*}
\mathbb{P}(\mathcal{G}(f) = I) &\in [0, 1]\\
\sum_{I \in S(\mathcal{G})} \mathbb{P}(\mathcal{G}(f) = I) &=1
\end{align*}

Latent factors's activations and signatures on the measure grid induce activations of measure grid's bits. We refer to this distribution over all possible combinations of activations of bits as $d$, and define it as

\begin{align*}
d &: F(2)^{n} \rightarrow [0,1]\\
d_x &= \mathbb{P}\left(b_1 = x_1, \ldots, b_n = x_n \right)
\end{align*}

Finally, we can also modelize the connection between factors and a measure grid's bit as a signature of the grid's bit on factor space. That is, for $I \in S(\mathcal{F})$, there is a well defined probability measure so that 

\begin{align*}
\mathbb{P}(\mathcal{G}^{-1}(b) = I) &\in [0, 1]\\
\sum_{I \in S(\mathcal{F})} \mathbb{P}(\mathcal{G}^{-1}(b) = I) &=1
\end{align*}

\subsubsection*{Characteristic polynome}

The activity of factors and grid's bits may be modelized using a set of multivariate polynomials whose fiber and image domain is respectively $F(2)^{n}$ and $F(2)$. The set of polynome associated with a set $I \in S(\mathcal{F})$ and $I' \in S(\mathcal{G})$ is denoted respectively $\lbrace \mathcal{P}_{I, l} \rbrace_{l \in \mathbb{N}}$  and $\lbrace \mathcal{P}_{I', l} \rbrace_{l \in \mathbb{N}}$. It represents a segmentation of states of respectively factors of $I$ and measure grid's bits of $I'$.

\begin{align*}
\mathcal{P}_{I, l}&: F(2)^{K} \rightarrow F(2)\\
\mathcal{P}_{I, l} [\textbf{x}] &= \begin{cases} \sum_{\pi \in S(I', l)} x_{\pi_1} \cdot \ldots \cdot x_{\pi_l}, & \text{if }\ l \in \lbrace 1, \ldots, \vert I \vert \rbrace \\ 0, & \text{otherwise} \end{cases}
\end{align*}

and 

\begin{align*}
\mathcal{P}_{I', l}&: F(2)^{n} \rightarrow F(2)\\
\mathcal{P}_{I', l} [\textbf{x}'] &= \begin{cases} \sum_{\pi \in S(I', l)} x'_{\pi_1} \cdot \ldots \cdot x'_{\pi_l}, & \text{if }\ l \in \lbrace 1, \ldots, \vert I \vert \rbrace \\ 0, & \text{otherwise} \end{cases}
\end{align*}

Where $S(I, l)$ and $S(I', l)$ are the set of all permutations of size $l$ of respectively $I$ and $I'$, $\textbf{x} = [x_{f_1}, \ldots, x_{f_K}]$ and $\textbf{x}' = [x_{b_1}, \ldots, x_{b_n}]$. Furthermore we define the characteristic polynomial of a set $I \in S(\mathcal{F})$ and $I' \in S(\mathcal{G})$ at level $l_0 \in \{0, \ldots, \vert I \vert \}$ and $l'_0 \in \{0, \ldots, \vert I' \vert \}$  as 

\vspace{10px}\noindent\begin{minipage}{.5\linewidth}
\begin{align*}
\mathcal{P}^{l_0}_{I} &: F(2)^{K} \rightarrow F(2)\\
\mathcal{P}^{l_0}_{I} &= \sum_{l = l_0}^{\vert I \vert} \mathcal{P}_{I, l}
\end{align*}
\end{minipage}%
\noindent\begin{minipage}{.5\linewidth}
\begin{align*}
\mathcal{P}^{l'_0}_{I'} &: F(2)^{n} \rightarrow F(2)\\
\mathcal{P}^{l'_0}_{I'} &= \sum_{l = l'_0}^{\vert I' \vert} \mathcal{P}_{I', l}
\end{align*}
\end{minipage}\vspace{15px}

So far, the addition is set to be the logical XOR in the definition of fields $F(2)$. However, in the rest of this report, we will use symbol $+$ and $\sum$ as representation of a logical OR in $F(2)$. This notation enables us to save a lot of time in writing complex polynomials. Denoting $\oplus$ the logical XOR and $\bar{x}$ the opposite of $x$, one have

\begin{equation*}
(x \cdot \bar{y}) \oplus (\bar{x} \cdot y) = x + y 
\end{equation*}  

\subsubsection*{Operators on polynome}

In order to qualify a set of factors and grid's bits, we define some basic operator. First, let  $I_{\mathcal{G}}$ be a subset of $S(\mathcal{G})$, we denote by $F_2(I_{\mathcal{G}}, \mathcal{G})$ the operator that transforms $I_{\mathcal{G}}$ into a set of $\vert I_{\mathcal{G}} \vert$ vector in $F(2)^{\vert \mathcal{G} \vert}$. 

\begin{equation*}
F_2(I_{\mathcal{G}}, \mathcal{G})  : S(\mathcal{G}) \rightarrow  F(2)^{\vert \mathcal{G} \vert \times \vert I_{\mathcal{G}} \vert} 
\end{equation*}

For each vector $X \in F_2(\{I\}, \mathcal{G})$ such that $I\in I_{\mathcal{G}}$, an entry takes value $1$ if the associated index belongs to $I$, 0 otherwise. This operator is convenient to evaluate the characteristic polynome. As an example, let $(I, I') \in S(\mathcal{G})^{2}$ and $l_0 \in \{ 1, \ldots \vert I \vert \}$ then 

\begin{align*}
\sum_{x \in F_2(\lbrace I' \rbrace, \mathcal{G})} \mathcal{P}_I^{l_0} \left[ x \right]  &= \begin{cases} 1, & \text{if }\ \vert I'\cap I \vert \geq l_0 \\ 0, & \text{otherwise} \end{cases}
\end{align*}

Furthermore, given the distribution over measure grid's bits activation $d$, we define the norm of a characteristic Polynome $\mathcal{P}_I^{l_0}$ with respect to $d$ as

\begin{align*}
\Vert . \Vert_{d} &: \mathcal{P}_{F_2(S(\mathcal{G}), \mathcal{G})} \rightarrow \left[ 0, 1\right] \\
\Vert \mathcal{P}_I^{l_0} \Vert_{d} &= \sum_{x \in F_2(S(\mathcal{G}), \mathcal{G})} \mathcal{P}_I^{l_0}\left[ x \right] \times d_x
\end{align*}

Where $\mathcal{P}_{F_2(S(\mathcal{G}), \mathcal{G})}$ the space of all polynomials with domain $F_2(S(\mathcal{G}), \mathcal{G})$ and $\times$ is the simple multiplication in $\mathbb{R}$. Finally, keeping previous notations, let $\lbrace \mathcal{P}_{I_{i}}^{l_i} \rbrace_{i=1, ..., k}$ a set of characteristic polynome for some integer $k \geq 2$, we define the product operator with respect to $d$ as 

\begin{align*}
\langle ., \ldots, . \rangle_{d} &: \mathcal{P}_{F_2(S(\mathcal{G}), \mathcal{G})}^{k} \rightarrow \left[ 0, 1\right] \\
\langle \mathcal{P}^{l_1}_{I_1}, \ldots,  \mathcal{P}^{l_k}_{I_k} \rangle_{d} &= \sum_{x \in F_2(S(\mathcal{G}), \mathcal{G})} \left( \mathcal{P}^{l_1}_{I_1} \left[ x \right] \cdot \ldots \cdot \mathcal{P}^{l_k}_{I_k} \left[ x \right] \right) \times d_x
\end{align*}

Where $\cdot$ denotes the usual multiplication in $F(2)$ and $\times$ is the simple multiplication in $\mathbb{R}$. Finally, each operator specified above can also be defined in the factor space, using the characteristic polynomial in factor space and the distribution over factors's activations $d'$.

\subsubsection*{Stochastic processus induced by factor's activation}

Factors's activations are observed as a strictly stationnary stochastic processus. That is for a couple $(I, l) \in \mathcal{F} \times \{1, \ldots, \vert I \vert\}$,  we associate a stochastic process $x_I^l[t]$ defined as

\begin{equation*}
x_I^l[t] = \begin{cases} 1, & \text{with probability }\ \Vert \mathcal{P}_I^l \Vert_{d'}\\ 0, & \text{Otherwise} \end{cases}
\end{equation*}

with $\{ x_I^l[t] \}_{t\in \mathbb{N}}$ 2-by-2 independant. Factors's signatures and their activations lead to bits's activations that are also observed as striclty stationary stochastic processus. Again for a couple $(I, l) \in \mathcal{G} \times \{1, \ldots, \vert I \vert\}$,  we associate a stochastic process $x_I^l[t]$ defined as

\begin{equation*}
x_I^l[t] = \begin{cases} 1, & \text{with probability }\ \Vert \mathcal{P}_I^l \Vert_{d} \\ 0, & \text{Otherwise} \end{cases}
\end{equation*}

with $\{ x_I^l[t] \}_{t\in \mathbb{N}}$ 2-by-2 independant.

\subsection{Firing Graph}
The firing graph is the main data structure used in our solution. In this section we propose a definition of it, as well as basic tools to support its analysis.
  
\subsubsection*{Graph specification}
The algorithm presented in this report use a particular data structure that we refer as firing graph and that we denote $G(V, D_w)$. 

\begin{itemize}
\item $V$ is the set of vertices $V = \lbrace v_1, \ldots, v_{\vert V \vert} \rbrace$
\item $D_w$ is the weighted direct link matrix, $D_w \in \mathbb{N}^{\vert V \vert \times \vert V \vert}$ and $\left[ D_w \right]_{i, j} = w$ indicate an edge of weight $w$ from vertex $v_i$ to vertex $v_j$ if $w > 0$  
\end{itemize}

$G$ is a directed weighted graph whose vertices are organized in layer. A vertex $v$ of some layer $i \in \mathbb{N}$ must have at least one incoming edge from a vertex of layer $i-1$. It may also have incoming edges from any vertices of layer $k \in \mathbb{N}, k < i$. Such a set of vertices will be referred as the input domain of $v$. Vertices of layer $0$ have empty input domains, they correspond to bits of the measure grid $\mathcal{G}$. Each vertex stores the tuple $(I, l_0)$

\begin{itemize}
\item $I$ the set of vertices at the tail of incoming edge of the vertex, referred as input set
\item $l_0$ the firing rate's lower bound of the vertex, referred as level, $l_0 \in \lbrace 1, \ldots, \vert I \vert \rbrace$ 
\end{itemize}

 \begin{figure}[H]
 \centering
   \includegraphics[scale=0.25]{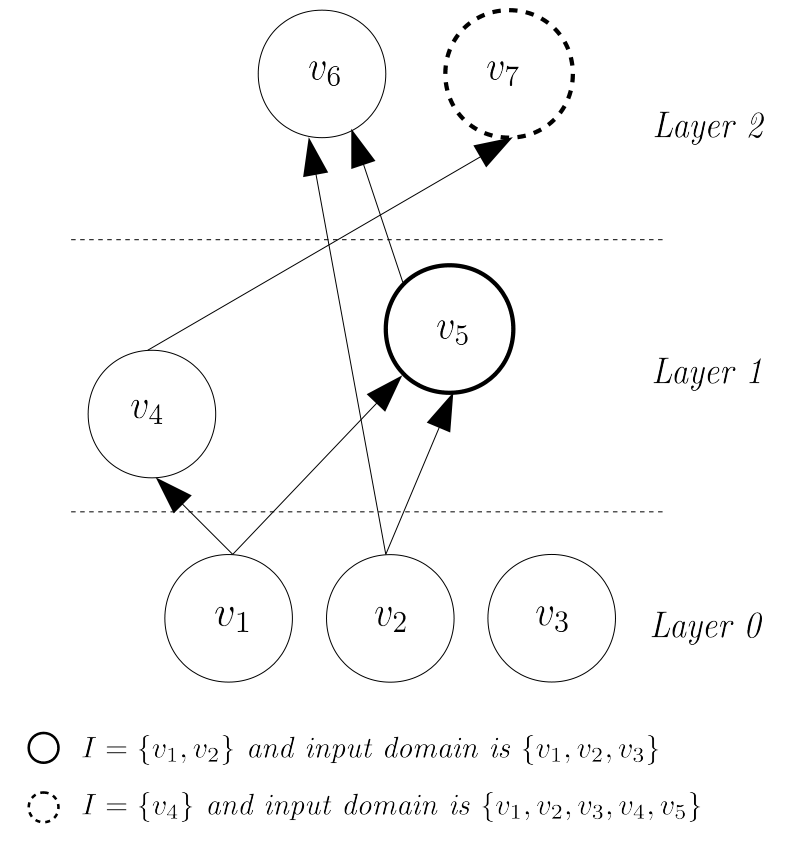}
 \caption{Firing graph}
 \label{fig:firing_graph}
\end{figure}

\subsubsection*{Graph Polynomials}

As for bits of the measure grid and factors, a vertex $v(I, l_0)$ of a firing graph is assiociated with the set of polynomes $\lbrace \mathcal{P}_{I, l} \rbrace_{l \in \{l_0, \ldots, \vert I \vert \ \}}$. Each polynome is a segment of its characteristic polynome $\mathcal{P}_v$ that describes activation, at instant $t$, of $v$, given its input domain's activations at instant $t-1$. If we denote by $n$, $I$ and $l_0$ respectively the size of the input domain of $v$, the set of vertex that has a link toward $v$ and the level of $v$, then 

\vspace{10px}\noindent\begin{minipage}{.5\linewidth}
\begin{align*}
\mathcal{P}_{v, l}&: F(2)^{n} \rightarrow F(2)\\
\mathcal{P}_{v, l}[\textbf{x}] &= \begin{cases} \sum_{\pi \in S(I, l)} x_{\pi_1} \cdot x_{\pi_2} \cdot \ldots \cdot x_{\pi_l}, & \text{if }\ l \in \lbrace l_0, \ldots, \vert I \vert \rbrace \\ 0, & \text{otherwise} \end{cases}
\end{align*}
\end{minipage}%
\noindent\begin{minipage}{.5\linewidth}
\begin{align*}
\mathcal{P}_v &: F(2)^{n} \rightarrow F(2)\\
\mathcal{P}_v[\textbf{x}]  &= \sum_{l = l_0}^{\vert I \vert} \mathcal{P}_{v, l}[\textbf{x}]
\end{align*}
\end{minipage}\vspace{15px}

Where $S(I, l)$ is the set of all permutations of size $l$ of elements of $I$ and $\textbf{x} \in F_2(\{ I \}, D_v)$, where $D_v$ is the input domain of $v$. Furthermore, all operators on polynome defined previously is applicable. Let $v, v_1, \ldots, v_k$  be some vertices of the firing graph with the same input domain and $d$ a distribution over activations of their input domain's vertices. Then the norm and the product with respect to distribution $d$ are defined as

\vspace{10px}\begin{minipage}{.3\linewidth}
\begin{align*}
\Vert \mathcal{P}_v \Vert_{d} = \sum_{x \in F_2(S(\mathcal{G}), \mathcal{G})} \mathcal{P}_v\left[ x \right] \times d_x
\end{align*}
\end{minipage}%
\noindent\begin{minipage}{.7\linewidth}
\begin{align*}
\langle \mathcal{P}_{v_1}, \ldots,  \mathcal{P}_{v_k} \rangle_{d} = \sum_{x \in F_2(S(\mathcal{G}), \mathcal{G})} \left( \mathcal{P}_{v_1} \left[ x \right] \cdot \ldots \cdot \mathcal{P}_{v_k} \left[ x \right] \right) \times d_x
\end{align*}
\end{minipage}\vspace{15px}

Finally, activations of vertices are observed as stochastic processus. Given a vertex $v(I, l)$ we define

\begin{equation*}
x_v[t] = \begin{cases} 1, & \text{with probability }\ \Vert \mathcal{P}_v \Vert_{d}\\ 0, & \text{Otherwise} \end{cases}
\end{equation*}

The stochastic process that takes value 1 if the vertex $v$ actvivates and 0 otherwise, at each instant of time. If measure grid's bits compound layer 0 of the firing graph, then, from definition of bit's stochastic processus and linearity of state's propagations, $x_v[t]$ is strictly stationary.

\subsubsection*{Connection to grid's bit}

The firing graph is a convenient data structure to measure activity of a complex group of measure grid's bits. When the firing graph's layer 0 is composed of measure grid's bits, the characteristic polynome of each vertex can be represented as a characteristic polynome in the measure grid's space, without consideration of time and delay. Let $G$ be such a firing graph, then for any vertex of layer 1, $v(I, \vert I \vert)$, the characteristic polynome $v$ is equal to the characteristic polynome of the set of bits $I \in \mathcal{G}$ with level $\vert I \vert$.

\begin{align*}
\mathcal{P}_{v} &= \mathcal{P}_{I, \vert I \vert}\\
x_v[t] &= x_{I}^{\vert I \vert}[t - 1] 
\end{align*}

Furthermore if we set the level of $v$ to 1 its characteristic polynome become the logical $or$-sum of the characteristic polynome of each bits of $I$

\begin{align*}
\mathcal{P}_{v} &= \sum_{b \in I} \mathcal{P}_{\{ b \}, 1}\\
x_v[t] &= \begin{cases} 1, & \text{if }\ \sum_{b \in I} x_{\{ b \}}^1[t - 1] > 0\\ 0, & \text{Otherwise} \end{cases}
\end{align*}

Besides, one can design more complexe arrangements of vertices that enable to model activations of multiple sets of measure grid's bits. Let $G$ be a firing graph with its layer 0 composed of $\mathcal{G}$, let $u(I, \vert I \vert)$ and $v(I', 1)$, such that $I \cap I' = \emptyset$, be vertices of layer 1 and $w(\{ u, v \}, 2)$ a vertex of layer 2. Then one can see that that characteristic polynome of $w$ verifies

\begin{align*}
\mathcal{P}_{w} &= \sum_{b \in I'} \mathcal{P}_{I \cup \{ b \}, \vert I \vert + 1}\\
x_w[t] &= \begin{cases} 1, & \text{if }\ \sum_{b \in I'} x_{I \cup \{ b \}}^{\vert I \vert +1}[t - 2] > 0\\ 0, & \text{Otherwise} \end{cases}
\end{align*}

\subsection{Evaluation of measure grid's bits}

A perfect indicator of the activation of a given factor $f$ can be used to evaluate the possibility of any set of bits to be part of $f$'s signature on the measure grid. 

\subsubsection*{Factor's signature}

One way to describe the activity of a factor $f$ on the measure grid is to associate it to a polynome in the measure grid's space

\begin{align*}
\mathcal{P}_{\mathcal{G}(f)} &: F(2)^n \rightarrow F(2)\\
 \mathcal{P}_{\mathcal{G}(f)} &=  \mathcal{P}_{\mathcal{G}(f), \vert \mathcal{G}(f) \vert}
\end{align*}

$\mathcal{P}_{\mathcal{G}(f)}$ is refered as the polynomial signature of $f$ on $\mathcal{G}$. Anytime $f$ is active then  its polynomial signature takes value 1. Yet under particular modelling of factor's links to measure grid, the polynomial signature of $f$ can take value 1 while $f$ is not active. More formally let $f \in \mathcal{F}$,  $\forall I \in S(\mathcal{F})$ such that $x \in F_2(\{I\}, \mathcal{F})$ and $x' \in F_2(\{\cup_{f \in I} \mathcal{G}(f) \}, \mathcal{G})$
\begin{center}
$\mathcal{P}_f[x] = 1 \Rightarrow \mathcal{P}_{\mathcal{G}(f)}[x'] = 1$
\end{center}

Furthermore if $!\exists J \in S(\mathcal{F} \setminus \{ f \})$ such that $\mathcal{G}(f) \subset \bigcup_{f' \in J} \mathcal{G}(f')$ then

\begin{center}
$\mathcal{P}_f[x] = 1 \Leftrightarrow \mathcal{P}_{\mathcal{G}(f)}[x'] = 1$
\end{center}

\subsubsection*{basic metrics}

Let $I \in S(\mathcal{G})$, $l \in \{ 1, \ldots, \vert I \vert \}$, $f \in \mathcal{F}$ and $e$ the event "factor $f$ is active". Then we define the recall coefficient of couple $(I, l)$ with respect to $f$ as

\begin{equation*}
\mu_{I, l, f} = \langle \mathcal{P}^{l}_{I}, \mathcal{P}_{\mathcal{G}(f)} \rangle_{d \vert e} + \langle \mathcal{P}^{l}_{I}, \bar{\mathcal{P}}_{\mathcal{G}(f)} \rangle_{d \vert e}
\end{equation*}

Where $d \vert e$ is the  distribution over bit's activations given event $e$ and $\bar{\mathcal{P}}_{\mathcal{G}(f)}$ is the complement of $\mathcal{P}_{\mathcal{G}(f)}$ in $F(2)$. Furthermore we define the precision coefficient of couple $(I, l)$ with respect to $f$ as

\begin{equation*}
\nu_{I, l, f} = \langle \mathcal{P}^{l}_{I}, \mathcal{P}_{\mathcal{G}(f)} \rangle_{d \vert \bar{e}} + \langle \mathcal{P}^{l}_{I}, \bar{\mathcal{P}}_{\mathcal{G}(f)} \rangle_{d \vert \bar{e}}
\end{equation*}

Where $d \vert \bar{e}$ is the  distribution over bit's activations given not event $e$. Finally we define the purity coefficient of couple $(I, l)$ with respect to $f$ as 

\begin{equation*}
\omega_{I, l, f} = \frac{\nu_{I, l, f}}{\mu_{I, l, f}}
\end{equation*}

The lower $\omega_{I, l, f}$ is, the purer is the couple ($I$, $l$) with respect to $f$. The recall, precision and purity coefficient can be defined for any vertex $v$ of a firing graph where vertices of layer 0 are composed by measure grid's bit and are denoted respectively $\mu_{v, f}$, $\nu_{v, f}$ and $\omega_{v, f}$. The latter are computed by using the representation of $\mathcal{P}_v$ as a characteristic polynomial in the measure grid's space.

\subsubsection*{advanced metrics}

Let $I \in S(\mathcal{G})$, $l \in \{ 1, \ldots, \vert I \vert \}$, $f \in \mathcal{F}$ and $e$ the event "factor $f$ is active". We define the precision of the couple $(I, l)$ with respect to factor $f$ as

\begin{equation*}
\phi_{I, l, f} = \frac{\Vert \mathcal{P}^{l}_I \Vert_{d, e}}{\Vert \mathcal{P}^{l}_I \Vert_{d}}
\end{equation*}

We also define the recall of the couple $(I, l)$ with respect to factor $f$ as

\begin{equation*}
\psi_{I, l, f} = \frac{\Vert \mathcal{P}^{l}_I \Vert_{d, e}}{\Vert \mathcal{P}_{\mathcal{G}(f)} \Vert_{d, e}}
\end{equation*}

Where $d, e$, the distribution over the combination of activations of measure grid's bits that intetersect with event $e$. The precision and the recall are defined for any vertex $v$ of a firing graph where vertices of layer 0 are composed by measure grid's bit and are denoted respectively $\phi_{v, f}$ and $\psi_{v, f}$. Again, The latter are computed by using the representation of $\mathcal{P}_v$ as a characteristic polynomial in the measure grid's space.

\subsubsection*{Advanced stochastic process induced by vertex}

Given a firing graph with its layer 0 composed of measure grid's bits, we have seen that the propagation of activations induces a stochastic process at each vertex. Here we introduce some more complex stochastic processus at each vertex of $G$. Given a vertex $v$ at layer $k \geq 0$, its  characteristic polynome $\mathcal{P}_{v}$, a factor $f \in \mathcal{F}$ and e, the event "factor $f$ is active", we define the score process of $v$ with respect to factor $f$ as

\begin{equation*}
s_{v,f}\left[N, T, p, q \right] = N + \sum_{t=1}^{T} s_{v, p, q, t, f}
\end{equation*}

Where $(N, T, p, q) \in \mathbb{N}^4$ and $\lbrace s_{v, p, q, t, f} \rbrace_{t \in \mathbb{N}}$ a set of i.i.d random variable. $s_{v, p, q, t, f}$ takes value $q$ if the event e was true at instant $t - k $ and value $-p$ if it was false, given that $v$ activates at instant $t$. That is, $\forall$ $t  < k $, $s_{v, p, q,t, f} = 0$ and $\forall$ $t  \geq k$

\begin{equation*}
s_{v, p,q, t, f} = \begin{cases} q, & \text{with probability } q_s  \\ -p, & \text{with probability } 1 - q_s \end{cases}
\end{equation*}

Where $q_s = \frac{q_r}{q_r + q_p}$ with $q_r = \Vert \mathcal{P}_v \Vert_{d,e} $ and $q_p = \Vert \mathcal{P}_v \Vert_{d} - q_r$.  $d, e$ is the distribution over measure grid's activations that intersect with the event e

\subsection{Properties}

This paragraph intend to deliver useful properties for the analysis of the algorithm. The proof of every properties can be found in the appendix A at the end of this paper.

\subsubsection*{Polynomial decomposition}

\underline{Partition}\\

Let $v_1(I, l_0)$, $v_2(J, 0)$ and $v_3(K, 0)$, be three vertices at the layer 1 of some firing graph, with the same input domain $\mathcal{G}$. If $I = J \cup K$ and $J \cap K = \emptyset$, then, $\forall x \in F_2(S(\mathcal{G}), \mathcal{G})$

\begin{equation}
\label{prop:partition-1}
\mathcal{P}_I^{l_0}\left[ x \right] = \sum_{l=l_0}^{\vert I \vert} \sum_{j=0}^{\vert J \vert} \mathcal{P}_{J, j}\left[ x \right]  \cdot \mathcal{P}_{K, l - j}\left[ x \right] 
\end{equation} 

In paticular for $b \in I$

\begin{equation}
\label{prop:partition-2}
\mathcal{P}_{I, l}\left[ x \right] = \mathcal{P}_{I\setminus \{ b\}, l}\left[ x \right] \cdot \mathcal{P}_{\{ b\}, 0}\left[ x \right] + \mathcal{P}_{I\setminus \{ b\}, l-1}\left[ x \right] \cdot \mathcal{P}_{\{ b\}, 1}\left[ x \right]
\end{equation} 

\underline{Decomposition}\\

Let $G$ be a firing graph with layer 0 composed of $\mathcal{G}$. Let $u(I, l_u)$, $v(I', l_v)$ such that $I \cap I' = \emptyset$ as vertices of layer 1 and $w(\{ u, v \}, 2)$ as vertex of layer 2. Let $K \in \cup_{l \in \{l_v, \ldots, \vert I' \vert \}} S(I', l)$, $x \in F_2(S(\mathcal{G}), \mathcal{G})$ and $x' = \begin{bmatrix} \mathcal{P}_{u}[x] &\mathcal{P}_{v}[x] \end{bmatrix}$ then

\begin{equation}
\label{prop:2layer-1}
\mathcal{P}_{K, \vert K \vert}\left[ x \right] \cdot \mathcal{P}_{\{u, v\}, 2}\left[ x' \right] = \sum_{l=l_u}^{\vert I \vert}  \sum_{J \in S(I, l)} \mathcal{P}_{J \cup K, l + \vert K \vert }\left[ x \right]
\end{equation}

In particular if $l_u = \vert I \vert$ and $l_v = 1$, then for any vertex of layer 0, $b \in I'$

\begin{equation}
\label{prop:2layer-2}
 \mathcal{P}_{b, 1}\left[ x \right] \cdot \mathcal{P}_{\{u, v\}, 2} \left[ x' \right]= \mathcal{P}_{I \cup \{b\}, \vert I \vert + 1}\left[ x \right]
\end{equation}

\subsubsection*{Metrics}

Throughout this section, we consider $G$ to be a firing graph with layer 0 composed by measure grid's bits $\mathcal{G}$ and $f \in \mathcal{F}$ denote some target factor that is linked to some bit of the measure grid. The distribution of activation of latent factors and measure grid's bits will be respectively denoted $d$ and $d'$ and e is the event "factor $f$ is active". Furthermore we use $v$ to denote some vertex of $G$ whose characteristic polynome respects $\mathcal{P}_v = \mathcal{P}_{I}^{l}$ with $(I, l) \in S(\mathcal{G}), \{1, \ldots, \vert I \vert\}$ and $f \in \mathcal{F}$ some factor\\

\underline{Precision of vertex}\\

The precision of $v$ with respect to $f$ is

\begin{equation}
\label{prop:precision1}
\phi_{v, f} = \frac{\Vert \mathcal{P}_{f} \Vert_{d'}}{\Vert \mathcal{P}_{f} \Vert_{d'} + (1 - \Vert \mathcal{P}_{f} \Vert_{d'}) \times \omega_{I, l, f}}
\end{equation}

Furthermore, if $\mu_{v, f} = 1$ we have

\begin{equation}
\label{prop:precision2}
\phi_{v, f} \leq \frac{\Vert \mathcal{P}_{f} \Vert_{d'}}{\Vert \mathcal{P}_{f} \Vert_{d'} + (1 - \Vert \mathcal{P}_{f} \Vert_{d'}) \times \omega_{\mathcal{G}(f),\vert \mathcal{G}(f) \vert, f}} 
\end{equation}

\underline{Recall of vertex}\\

The recall of $v$ with respect to $f$ is

\begin{equation}
\label{prop:recall1}
\psi_{v, f} = \mu_{I, l, f}
\end{equation}

Furthermore, 

\begin{equation}
\label{prop:recall2}
0 \leq \phi_{v, f} \leq 1
\end{equation}

Where right equality is reached whenever $v$ is connected to a set of measure grid's bit $I \in \mathcal{G}$, with level $l_0 = \vert I \vert$ such that $I \subset \mathcal{G}(f)$. \\

\underline{vertex's score process}\\

If $s_{v, f}[N, T, p, q]$ denotes the score process of $v$ with respect to $f$, with $N, T, p, q \in \mathbb{N}^4$, then

\begin{equation}
\label{prop:score_mean}
\mathbb{E} \left[ s_{v, f}[N, T, p, q]  \right] = N + T \times (\phi_{I, l, f} \times (p + q) - p)
\end{equation}

Furthermore,

\begin{equation}
\label{prop:score_var}
\Var \left[ s_{v, p, q, t, f}  \right] = (q + p)^{2} \times \phi_{I, l, f} \times (1 - \phi_{I, l, f})
\end{equation}

\section{Identification of Latent Factor}
In this section, we present a procedure to identify a latent factor's activation. The procedure consists of two steps:

\begin{itemize}
\item Sampling: Sample the measure grid and build a firing graph.
\item Draining: Drain the firing graph to exclude high purity coefficient's vertices.
\end{itemize}

Both processus will be described and the efficiency of the draining algorithm quantified.

\subsection{Sampling}

Sampling the measure grid consists in following a procedure to select some bits of it. This procedure is usually designed to be the most efficient in the fullfilment of specific quantitative objective. First, we assume that we have access to a determinist exact indicator of $f$'s activations with $f\in \mathcal{F}$. Then, the objective of sampling is to maximize the probability that we sample a bit whose purity coefficient with respect to $f$ is lower or equal to some positive constant $\omega$. That is, if we denote $s$ the random variable of the outcome of a single sampling, the objective is to maximize

\begin{center}
$\mathbb{P}( \omega_{\{s\}, 1, f} \leq \omega)$
\end{center}  

Again, if we have a set $I \subset \mathcal{G}$ of bits, the objective of sampling is to maximize the probability of selecting a bit $b$, for which the purity of $I \cup \{b \}$ at level $\vert I \vert + 1$ is lower to a given positive constant $\omega$. That is, if we denote by $s$ the random variable of the outcome of a single sampling, the objective is to maximize

\begin{center}
$\mathbb{P}( \omega_{I \cup \{ s\}, \vert I \vert, f} \leq \omega)$
\end{center} 

We propose a very intuitive sampling method based on the indicator of activation of target factor $f$. Given parameters $p_s \in [0, 1]$ and $S_p$ respectively the probability of picking a bit and a set of pre-selected measure grid's bits, the sampling procedure writes

\begin{algorithm}[H]
\caption{Sampling}
\textbf{Input:} $p_{\mathcal{S}}$, $S_{p}$\\
\textbf{Output:} $S$
\begin{algorithmic}
\State $S \gets \{ \}$, $x_f \gets nextFactorState()$, $X_{\mathcal{G}} \gets nextGridState()$
\While{$ S \textit{ is empty}$}
 \If {$x_f = 1$ and $\forall b \in S_p \textit{ } X_{\mathcal{G}}[b] = 1$}
  \ForAll{$b \in \mathcal{G}\setminus S \cup S_p$}
   \If {$X_{\mathcal{G}}[b] = 1$}
    \State $S \gets S \cup \{ b \} \textit{ with probability } p_s$
   \EndIf	
  \EndFor	
 \EndIf
 \State $x_f \gets nextFactorState()$
 \State $X_{\mathcal{G}} \gets nextGridState()$
\EndWhile
\end{algorithmic}
\end{algorithm}

Where $x_f$ and $X_{\mathcal{G}}$ are respectively a scalar that takes value 1 when factor $f$ is active, 0 otherwise, and a mapping with measure grid's bits as keys and their states as values (0 or 1). The second mean of the sampling procedure is to build a firing graph. The construction of the firing graph requires to set a parameter $N \in \mathbb{N}$ that corresponds to the initial weigth of edges that will be drained. In addition we set a mask matrix $G_{mask} \in \{0, 1\}^{\vert V \vert}$ that controls which vertex is allowed to have their outcoming edges updated during draining. We consider two kind of firing graphs.

\begin{figure}[H]
\centering
\includegraphics[scale=0.35]{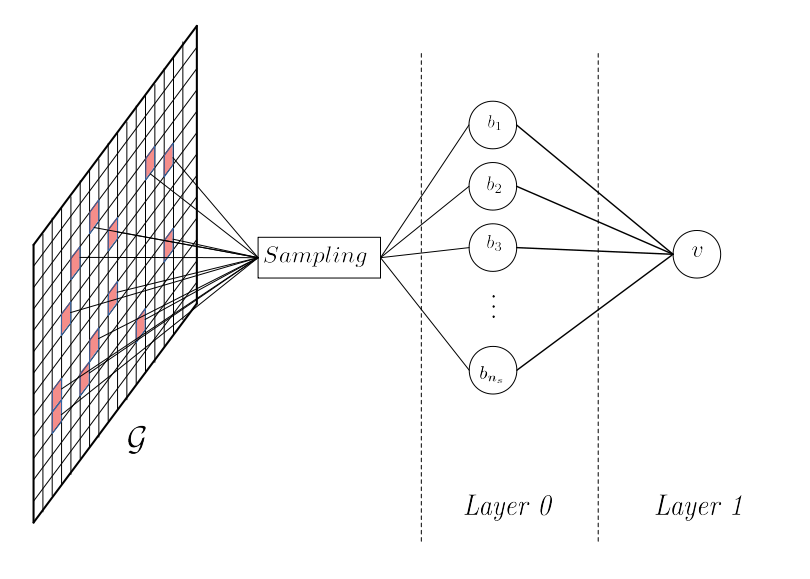}
\caption{Single sampled firing graph}
\label{fig:single_sampled_fg}
\end{figure}

In figure~\ref{fig:single_sampled_fg}, sampled bits $\{b_1, \ldots, b_{n_s}\}$ are used as vertices of the layer 0 of a firing graph $G$, $n_S = \vert S \vert$. Then vertex $v(\{b_1, \ldots, b_{n_s}\}, 1)$ is added at the layer 1 of $G$. Furthermore, we set $G_{mask}$ so to allow only layer 0's outcoming edges to be updated through draining.

\begin{figure}[H]
\centering
\includegraphics[scale=0.35]{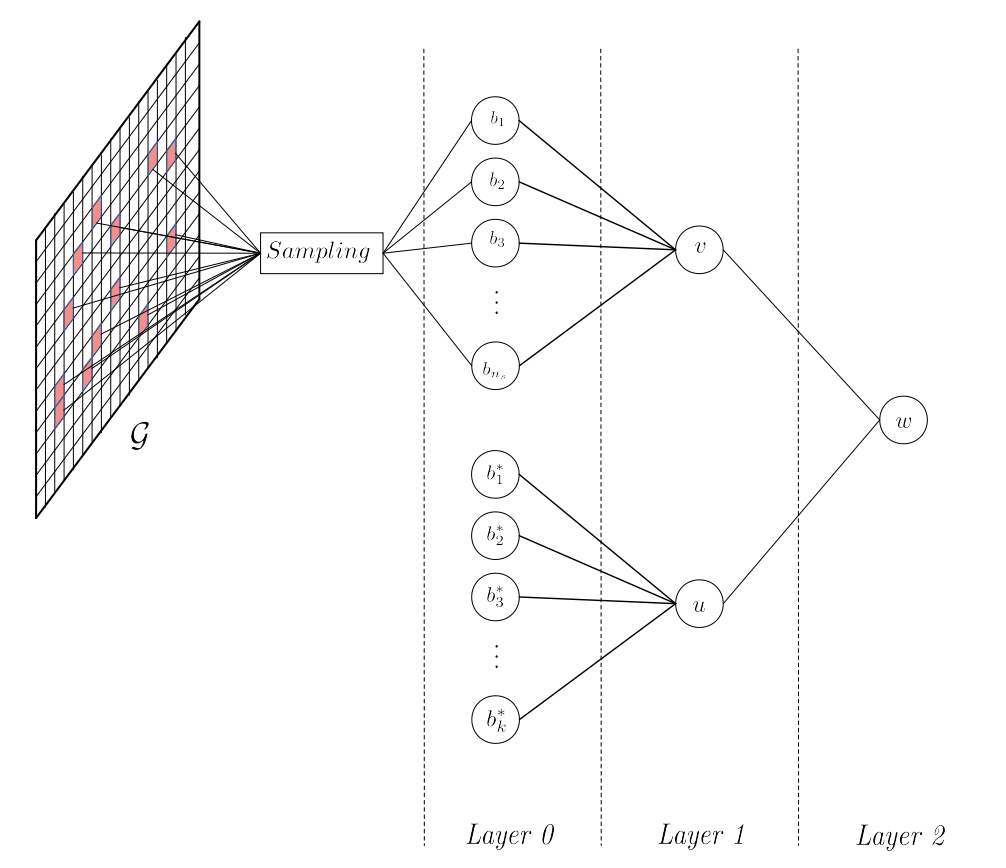}
\caption{Joint sampled firing graph}
\label{fig:joint_sampled_fg}
\end{figure}

In figure~\ref{fig:joint_sampled_fg}, sampled bits $\{b_1, \ldots, b_{n_s}\}$ and pre-selected bits $\{b_1^{*}, \ldots, b_k^{*} \}$ for some $k \in \mathbb{N}^{*}$ compound the layer 0 of the firing graph $G$, $n_S = \vert S \vert$. Then, vertices $v(\{b_1, \ldots, b_{n_s}\}, 1)$ and $u(\{b_1, \ldots, b_{k}\}, k)$ are added at layer 1 of $G$ and vertex $w(\{u, v\}, 2)$ at layer 2 of $G$. Finally, we set $G_{mask}$ so that only $b_1, \ldots, b_{n_s}$'s outcoming edges are allowed to be updated through  draining.

\subsection{Draining}

Draining the firing graph consists in iterating a forward propagation of bits's activations and a backward propagation of feedback generated by factor's activations through the firing graph. Feedback are meant to increment or decrement the weight of unmasked vertices's outcoming edges. Given that an edge with a null or negative weigth vanishes, at the end of the routine, connections of the graph differentiate between vertices's purity. To ease understanding of the algorithm, we split vertices of the firing graph into input and core vertices which are respectively vertices of layer 0 and vertices of layers $> 0$. Furthermore, we introduce a new type of vertices that can only have incoming edges from core vertices. We refer to those vertices as outputs. 

\begin{figure}[H]
\centering
\includegraphics[scale=0.30]{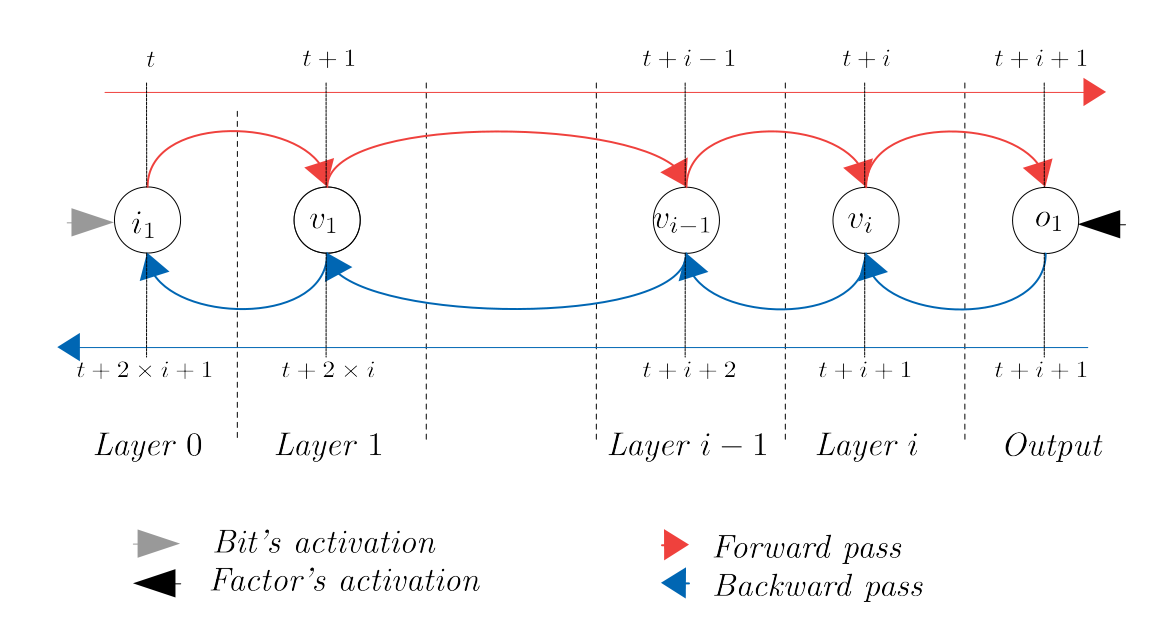}
\caption{Draining diagram}
\label{fig:draining_diagram}   
\end{figure}

We use $n_i$, $n_c$ and $n_o$ to refer to the number of respectively input, core and output vertices. Furthermore, we define $I_w \in \mathbb{N}^{n_i \times n_c}$, $C_w \in \mathbb{N}^{n_c \times n_c}$ and $O_w \in \mathbb{N}^{n_c \times n_o}$ that correspond to the weighted direct link matrices respectively from input toward core vertices, core toward core vertices and core toward output vertices. Furthermore we will use $A = A_w > 0$, $A \in \{I, C, O \}$ to denote the corresponding unweighted direct link matrices. Finally, in order to represent in a more convenient way stochastic processus induced by measure grid's activations, we define the following stochastics vectors

\begin{itemize}
\item $x_i^{(t)} \in \{0, 1\}^{1 \times n_i}$ the vector of activations of input vertices at instant $t$
\item $x_c^{(t)} \in \{0, 1\}^{1 \times n_c}$ the vector of activations of core vertices at instant $t$
\item $x_o^{(t)} \in \{0, 1\}^{1 \times n_o}$ the vector of activations of output vertices at instant $t$

\end{itemize}

The propagation of activations through the firing graph can be represented with two equations:\\

\vspace{20px}\noindent\begin{minipage}{.5\linewidth}
\begin{center}
\underline{Forward transmitting (FT)}
\end{center}
\begin{align*}
\tilde{x}_c^{(t)} &= x_i^{(t-1)} \cdot I + x_c^{(t-1)} \cdot C  \\
\tilde{x}_o^{(t)} &= x_c^{(t-1)} \cdot O
\end{align*} 
\end{minipage}%
\noindent\begin{minipage}{.5\linewidth}
\begin{center}
\underline{Forward processing (FP)}
\end{center}
\begin{align*}
[x_c^{(t)}]_i &= \begin{cases} 1, & \text{if }\ [\tilde{x}_c^{(t)}]_i > l_i \\ 0, & \text{Otherwise} \end{cases}\\
[x_o^{(t)}]_j &= \begin{cases} 1, & \text{if }\ [\tilde{x}_o^{(t)}]_j > 1 \\ 0, & \text{Otherwise} \end{cases}  
\end{align*}
\end{minipage}\vspace{20px}

Where $\cdot$ is the usual matrix multiplication, $(i, j) \in \{1, \ldots n_c\} \times \{1, \ldots n_o\}$ and $l_i$ is the level of the $i^{th}$ core vertex. An output vertex of the firing graph is fed with the activation of a targeted factor decayed in time by the number of layer - 1. That is, for single and joint sampled firing graphs, the decay is respectively set to 1 and 2. Factor's activations generate a feedback to the output that is back propagated through the firing graph. Supposing that we set the factor's decay to $d \geq 1$, the feedback is defined as 

\begin{equation*}
x_{b,o}^{(t)} = x_o^{(t)} \circ \left( (p + q) \times x_f^{(t-d)} - p \right)
\end{equation*}

Where $\circ$ denotes the Hadamard product, $x_f^{(t-d)}$ is the vector of states of factors at instant $t-d$ and $p, q$ are pre-difined positive integers. A correct backpropagation of $x_{b,o}^{(t)}$ up to the input vertices is made possible by using time and space coherence of firing graph's forward states. We denote by $V_i$, $i\in \mathbb{N}$ the set of vertices that has a path, composed of $i$ vertices, toward an output vertex. Let $G$ be a firing graph with $k \in \mathbb{N}^{*}$ layers augmented with a layer of ouptut vertices. Let $V_o$ the set of output vertices, $\forall v, o \in V_0 \times V_o$,  $v$ is elligible to $o$'s feedback at instant $t$ if and only if 

\begin{itemize}
\item $v$ was active at instant $t-1$
\item $v$ has an edge toward $o$
\end{itemize}    

The same principle can be used to backpropagate the feedback from vetices of $V_0$ towards vertices of $V_1$ and so on. Generally speaking, the back propagation from vertices of $V_i$ towards $V_{i+1}$ respects  $\forall v, v' \in V_{i} \times V_{i-1}$,  $v$ is elligible to feedback of $v'$ at instant $t$ if and only if 

\begin{itemize}
\item $v$ was active at instant $t - (2 \times i + 1)$
\item $v$ has an edge toward $v'$
\end{itemize}    

Finally we can encode the backpropagation equations as

\vspace{20px}\noindent\begin{minipage}{.5\linewidth}
\begin{center}
\underline{Backward transmitting (BT)}
\end{center}
\begin{align*}
\tilde{X}_{b, c}^{(t)} &= (O \cdot X_{b, o}^{(t-1)} + C \cdot X_{b, c}^{(t-1)}) \circ X_{m, c}^{(t)T}\\
X_{b,i}^{(t)} &= (I \cdot X_{b, c}^{(t-1)}) \circ X_{m, i}^{(t)T}
\end{align*} 
\end{minipage}%
\noindent\begin{minipage}{.5\linewidth}
\begin{center}
\underline{Backward processing (BP)}
\end{center}
\begin{align*}
X_{b, o}^{(t)} &= \begin{bmatrix} \textbf{0}_{no \times 1} & x_{b, o}^{(t)} & \textbf{0}_{no \times d_{max} - 2} \end{bmatrix} \\
X_{b, c}^{(t)} &= \begin{bmatrix} \textbf{0}_{n_c \times 2} & [\tilde{X}_{b, c}^{(t)}]_{(:n_c, :(d_{max}-2))} \end{bmatrix} 
\end{align*}
\end{minipage}

\begin{center}
\underline{Structure udpates (SU)}
\begin{align*}
O_w &= O_w + O \circ (X_{b, o}^{(t-1)} \cdot X_{m, c}^{(t)})^{T}\\
C_w &= C_w + C \circ (X_{b, c}^{(t-1)} \cdot X_{m, c}^{(t)})^{T}\\
I_w &= I_w + I \circ (X_{b, c}^{(t-1)} \cdot X_{m, i}^{(t)})^{T}
\end{align*}
\end{center}\vspace{20px}

Where $X_{m,c}^{(t)} = \begin{bmatrix} x_c^{(t)} & \ldots & x_c^{(t- d_{max})}\end{bmatrix}^{T}$ and $X_{m, i}^{(t)} =\begin{bmatrix} x_i^{(t)} & \ldots & x_i^{(t- d_{max})}\end{bmatrix}^{T}$, $X_{c, b}^{(t)} \in  \{0, q, - p \}^{n_c \times d_{max}}$ for $t \in \mathbb{N}^{*}$ and $X_{c, b}^{(0)} = \textbf{0}_{n_c \times d_{max}}$. Furthermore $d_{max} \geq (l -1) \times 2 + 1$ where $l$ is the number of layers of the firing graph. Finally we provide a parameter $T \in \mathbb{N}$ to the draining algorithm. It controls the targeted number of feedback that an edge should receive before disabling its update. Maintaining update's permissions for each edge requires an operation similar to structure updates. Finally, the draining algorithm iterates forward and backward pass until either $G$ is composed of two distinct connexe components, no structure update is enabled or the maximum number of iterations $T_{max} \in \mathbb{N}$ has been reached. 

\begin{algorithm}[H]
\caption{Draining}
\textbf{Input:} $G$, T, $T_{max}$, $p$, $q$, decay\\
\textbf{Output:} $G$ drained
\begin{algorithmic}
\State $i \gets 0$\Comment{Initialisation}
\State $X_{b,c}$, $x_{b, o}, x_i, x_c, X_{m, c}, X_{m, i}$ $\gets$ InitSignals()
\While{$i < T_{max}$}\Comment{Core loop}
 \State $x_i \gets nextGridState()$
 \State $x_c, x_o \gets \textit{FT}(G, x_i, x_c)$\Comment{Forward pass}
 \State $X_{m, c}, X_{m, i}, x_c, x_o \gets \textit{FP}(x_c, x_o)$
 \If {$i \geq decay$}
  \State $x_f \gets nextFactoreState()$
  \State $X_{b,c}, X_{b, o} \gets \textit{BP}(X_{b, c}, X_{b, o} x_f, p, q)$\Comment{Backward pass}
  \State $G' \gets \textit{SU}(T, G, X_{b, c}, X_{b, o}, X_{m, c}, X_{m, i})$
  \State $X_{b, c}, X_{b, i} \gets \textit{BT}(G, X_{b, c}, X_{b, o}, X_{m, c}, X_{m, i}, )$
  \State $G \gets G'$
 \EndIf	
 \If {$G.cc == 2$ or $\textit{not } G_{mask}.any()$}\Comment{Stop conditions}
  \State $break$
 \EndIf	
 \State $i \gets i + 1$ 
\EndWhile
\end{algorithmic}
\end{algorithm}

Clearly, the complexity of the algorithm is dominated by the backward transmit and structure updates operations. A standard worst case analysis of those operations gives $\mathcal{O}(n^{4} \times d_{max}^2)$, where $n$ is the total number of vertices in the firing graph. Yet this analysis relies on standard complexity time for dense matrix operations, and does not take into account neither the sparsity of signals and direct link matrices nor the distribution of input vertices's activations. In practice, we have found that the forward and backward propagation of bits and factors's activations is time consuming, especially when both $N$ and $T$ are large numbers. Thus, to reduce running time, batch\_size successive bits and factors's states are forward and backward propagated with an efficient vectorization of the equation. The decrease in time complexity of this practical trick is impressive and worth the gain in space complexity of the algorithm. Finally this trick may requires to dynamically change the batch\_size so that treshold for the number of updates at each edges is respected.

\subsection{Analysis of the algorithm}

\begin{theorem}
\label{th_stopping}
Given a set of sampled bits $S$, a set of pre-selected bits $I =\{b_{1}^{*}, \ldots, b_{i}^{*}\}$ a target factor $f$ and $G$, the firing graph built after sampling algorithm. A 5-tuple $(\omega, N, T, p, q)$ exists such that the probability of event E: "no input vertices of $G$ have outcoming edges at the end of the draining" is upper bounded. More specifically

\begin{equation*}
\mathbb{P} \left( E \right) \leq \sum_{j = 0}^{\vert S \vert} p_{-}^j \times \mathbb{P}_{\mathcal{S}} \left( \vert \lbrace s \in S \setminus \omega_{I \cup \{s\}, \vert I \vert + 1, f } < \omega \rbrace \vert = j \right) 
\end{equation*}

Where $p_{-} = \mathbb{P} \left( s_{v, f}[N, T, p,q] < 0 \vert \omega_{I \cup \{ s \}, i + 1, f} < \omega \right)$. Where $v(I \cup \{ s \}, i+1)$, for any $s \in S$, is a vertex of layer 1 of a firing graph $G$ of 2 layers. Furthermore

\begin{equation*}
\mathbb{P} \left( s_{v, f}[N, T, p, q]< 0 \vert \omega_{I \cup \{ s \}, i + 1, f} < \omega \right) \leq C \times \max \left( \exp\left(- T \times \left(\frac{\delta_{f} c}{\sigma}\right)^2 \right), \exp\left(- T \times \delta_{f}c \right) \right)
\end{equation*}

With $\delta_{f}$, $C$ and $c$ are postitive constants that depends on $\omega$ and $i$ and $f$. $\Var[s_{v, p, q, f, t}] = \sigma^2$. 

\end{theorem}

\textbf{Proof.} As a reminder, in the core of this proof, we refer to $d$ and $d'$ respectively to the distribution over bits's activations and factors's activations. Given the arrangement of vertices of graph $G$ and the forward equations of the draining algorithm, the activation of any vertices $b \in S$ that will be propagated toward an output vertex, is modelled by the following characteristic polynomial

\begin{equation*}
\mathcal{P}_{\{b\}, 1} \cdot \mathcal{P}_{\{u, v\}, 2}
\end{equation*}

With $v(S, 1)$ and $u(\{b_1^{*}, \ldots, b_i^{*}\}, i)$. Thus, using (\ref{prop:2layer-2}), the activity of $b$ that is propagated to the ouptut vertex is the same than the activity of a vertex $v(\{b_1^{*}, \ldots, b_i^{*}, b\}, i+1)$ at the layer 1 of a firing graph $G'$ where $b_1^{*}, \ldots, b_i^{*}$ and $b$ compound its layer 0. Furtermore, given the time and space consistency of the backpropagation of the feedback from the output vertex, the weight of the outcoming edge of $b$, at the convergence of the draining algorithm, is either $0$ or equal to the score process of vertex $v$ in $G'$ with respect to $f$, $s_{v, f}[N, T, p, q]$. Then, the first inequality is obtained by developping

\begin{align*}
\mathbb{P} \left( E \right)&= \sum_{j=0}^{\vert S \vert} p_{-}^j \times p_{+}^{\vert S \vert -j} \times\mathbb{P}_{\mathcal{S}} \left( \vert \lbrace s \in S \setminus \omega_{I \cup \{s\}, i + 1, f } < \omega \rbrace \vert = j \right) \\
&\leq \sum_{j=0}^{\vert S \vert} p_{-}^j \times \mathbb{P}_{\mathcal{S}} \left( \vert \lbrace s \in S \setminus \omega_{I \cup \{s\}, i + 1, f } < \omega \rbrace \vert = j \right)
\end{align*}  

Where 
\begin{itemize}
\item $I =\{b_{1}^{*}, \ldots, b_{i}^{*}\}$
\item $p_{-} = \mathbb{P} \left( s_{v, f}(N, T, p, q) < 0 \vert \omega_{I\cup\{s \}, i + 1, f} < \omega \right)$
\item $p_{+} = \mathbb{P} \left( s_{v, f}(N, T, p, q) < 0 \vert \omega_{I\cup\{s \}, i + 1, f} \geq \omega \right)$
\end{itemize}

Then, we choose the value of the postive real $\omega$ such that a measure grid's bit $b^{+}$ verifies

\begin{align*}
b^{+} = &\argmin_{b \in \mathcal{G}} \vert \omega - \omega_{I \cup \{ b \}, i +1, f} \vert \\
&\textit{ such that } \omega - \omega_{I \cup \{ b \}, i +1, f} > 0 
\end{align*}

And we define the vertex $v^{+}(I \cup \{b^{+}\}, i+1)$ and $\delta^{+} = \vert \omega - \omega_{v^{+}, f} \vert$. If vertex $v$ is such that $\omega_{v,f}  < \omega$ then using (\ref{prop:precision1}) one gets

\begin{equation*}
\underbrace{\frac{\Vert \mathcal{P}_{f} \Vert_{d'}}{\Vert \mathcal{P}_{f} \Vert_{d'} + (\omega - \delta) \times (1 - \Vert \mathcal{P}_{f} \Vert_{d'})}}_{\phi_{v, f}} \geq \underbrace{\frac{\Vert \mathcal{P}_{f} \Vert_{d'}}{\Vert \mathcal{P}_{f} \Vert_{d'} + (\omega - \delta^{+}) \times (1 - \Vert \mathcal{P}_{f} \Vert_{d'})}}_{\phi_{v^{+}, f}} > \underbrace{\frac{\Vert \mathcal{P}_{f} \Vert_{d'}}{\Vert \mathcal{P}_{f} \Vert_{d'} + \omega  \times (1 - \Vert \mathcal{P}_{f} \Vert_{d'})}}_{\phi}
\end{equation*}

For some real $\delta \geq \delta^{+} > 0$. Then, We choose the 4-tuple $(N, T, p, q)$ as follow:

\begin{align*}
(p, q) &\in \mathbb{N}^{2} \textit{ such that } \phi \times (p + q) - p < 0\\
N &= -T \times (\phi \times (p+q) - p)\\
T &\in \mathbb{N} \textit{ such that } N \textit{ large enough}
\end{align*}

Thus, given $\omega_{v,f}  < \omega$ one can write

\begin{align*}
\mathbb{P} \left( s_{v,f}[N, T, p, q] < 0 \right) &= \mathbb{P} \left( N + \sum_{t=1}^{T} s_{v, p, q, T, f} < 0 \right) \\
&= \mathbb{P}\left( \sum_{t=1}^{T} s_{v, p, q, t, f} - T \times \mathbb{E}\left[s_{v, p, q, 1, f}\right] < -N -  T \times \mathbb{E}\left[s_{v, p, q, 1, f}\right]\right)
\end{align*}

Furthermore from the definition of $\phi$ and $\phi_{v, f}$ we have

\begin{equation*}
\phi_{v, f} =\phi +  \underbrace{\delta \times \phi \times \phi_{v, f} \times \frac{1 - \Vert \mathcal{P}_{f} \Vert_{d'}}{\Vert \mathcal{P}_{f} \Vert_{d'}}}_{\delta_{v, f}}
\end{equation*}

Yet using equation (\ref{prop:score_mean}) one have

\begin{align*}
\mathbb{E}\left[s_{v, p, q, 1, f}\right] &= \phi_{v, f} \times (p + q) - p\\
&= (\phi + \delta_{v, f}) \times (p + q) - p\\
\end{align*}

Using $N = - T \times \left(\phi \times (p + q) - p \right)$ and the definition of $\phi$ one have

\begin{equation*}
-N -  T \times \mathbb{E}\left[s_{v, p, q, 1, f}\right] = -T \times (p + q) \times \delta_{v, f} 
\end{equation*}

Thus

\begin{align*}
\mathbb{P} \left( s_{v,f}[N, T, p, q] < 0 \right) &= \mathbb{P}\left( \sum_{t=1}^{T} s_{v, p, q, t, f} - \mathbb{E}\left[s_{v, p, q, t, f}\right] < -T \times (p + q) \times  \delta_{v, f} \right)\\
&\leq \mathbb{P}\left(\vert \sum_{t=1}^{T} s_{v, p, q,t, f} - \mathbb{E}\left[s_{v, p, q, t, f}\right] \vert > T \times (p + q) \times  \delta_{v, f} \right)\\
&\leq \mathbb{P}\left(\vert \sum_{t=1}^{T} s_{v, p, q, t, f} - \mathbb{E}\left[s_{v, p, q, t, f}\right] \vert > T \times \delta_{f} \right)
\end{align*}

With $\delta_{f} = (p + q) \times \delta^{+} \times \phi^{2} \times \frac{1 - \Vert \mathcal{P}_{f} \Vert_{d'}}{\Vert \mathcal{P}_{f} \Vert_{d'}}$. \\

At this point we have to notice that $\lbrace s_{v, p, q, t, f} \rbrace_{t=1, \ldots, T}$ is a sequence of i.i.d random variables with mean $\mu$ and variance $\sigma^2$ that verifies $\vert s_{v, p, q, t, f} \vert \leq max(p, q)$. Thus one can apply the Chernoff inequality as formulated in \cite{TAO-1}. In particular, taking $\lambda = \sigma^{-1}\delta_{f}$ we obtain

\begin{align*}
\mathbb{P}\left(\vert \sum_{t=1}^{T} s_{v, p, q, tn f} - \mathbb{E}\left[s_{v, p, q, t, f}\right] \vert > T \delta_{f} \right) &= \mathbb{P}\left(\vert \sum_{t=1}^{T} s_{v, p, q, t} - \mathbb{E}\left[s_{v, p, q, t}\right] \vert > \lambda\sigma \sqrt{T} \right)\\
&\leq C \times \max \left( \exp\left(- T \times \left(\frac{\delta_{f} c}{\sigma}\right)^2 \right), \exp\left(- T \times \delta_{f}c \right) \right)
\end{align*}

With $C, c$ some positive constant and $\Var[s_{v, p, q, t, f}] = \sigma^2$ for $t \in \{1, \ldots, T \}$. Q.E.D.

\begin{center}
\rule[0pt]{100pt}{1pt} 
\end{center}

\begin{theorem}
\label{th_precision}
Given a set of sampled bits $S$, a set of pre-selected bits $I =\{b_{1}^{*}, \ldots, b_{i}^{*}\}$ a target factor $f$ and $G$ the firing graph built after sampling algorithm. A sequence of 5-tuple $(\omega, N, T, p, q)$ exists such that for each input vertex $v$ of $G$, from which the output is reachable, we have 

\begin{equation}
\mathbb{P} \left( \omega_{v, f} > \omega \right) \leq C \times \max \left( \exp\left(- T \times \left(\frac{\delta_{f} c}{\sigma}\right)^2 \right), \exp\left(- T \times \delta_{f}c \right) \right)
\end{equation}

Where $v(I \cup \{ s \}, i+1)$, for any $s \in S$, is a vertex of layer 1 of a firing graph $G$ of 2 layers and $\delta_{f}$, $C$ and $c$ are postitive constants that depends on $\omega$ and $i$ and $\Var[s_{v, p, q, f, t}] = \sigma^2$. 
\end{theorem}

\textbf{Proof.} 
As in the proof of the previous theorem, using the arrangement of vertices of $G$, the property (\ref{prop:2layer-2}) and the forward and backward equations of the draining algorithm, one can show that the weight of the outcoming edge of any vertices $b \in S$ of $G$ is either equal to 0 or to the score process $s_{v, f}[N, T, p, q]$ where $v(\{b_1^{*}, \ldots, b_i^{*}, b\}, i+1)$ is a vertex at the layer 1 of a firing graph $G'$ where $b_1^{*}, \ldots, b_i^{*}$ and $b$ compound its layer 0. Furthermore, if sample $b$ still have outcoming edges after draining, then

\begin{equation*}
\mathbb{P} \left( \omega_{v, f} >  \omega \right) = \mathbb{P}_{\mathcal{S}} \left( \omega_{I \cup \{b\}, i + 1, f } > \omega \right) \times \mathbb{P} \left( s_{v, f}(N, T, p, q) > 0 \vert \omega_{I \cup \{ b \}, i + 1, f} > \omega \right)
\end{equation*}

Then, we choose the value of the postive real $\omega$ such that a bit $b^{-}$ verifies

\begin{align*}
b^{-} = &\argmin_{b \in \mathcal{G}} \vert \omega - \omega_{I \cup \{ b \}, f} \vert \\
&\textit{ such that } \omega - \omega_{I \cup \{ b \}, f} < 0 
\end{align*}

And we define the vertex $v^{-}(I \cup \{b^{-}\}, i+1)$ and $\delta^{-} = \vert \omega - \omega_{v^{-}, f} \vert$. If $v$ is such that $\omega_{v, f}  < \omega$ then using (\ref{prop:precision1}) we have 

\begin{equation*}
\underbrace{\frac{\Vert \mathcal{P}_{f} \Vert_{d'}}{\Vert \mathcal{P}_{f} \Vert_{d'} + (\omega + \delta) \times (1-\Vert \mathcal{P}_{f} \Vert_{d'})}}_{\phi_{v, f}} \leq \underbrace{\frac{\Vert \mathcal{P}_{f} \Vert_{d'}}{\Vert \mathcal{P}_{f} \Vert_{d'} + (\omega + \delta^{-}) \times (1 - \Vert \mathcal{P}_{f} \Vert_{d'})}}_{\phi_{v^{-}, f}} < \underbrace{\frac{\Vert \mathcal{P}_{f} \Vert_{d'}}{\Vert \mathcal{P}_{f} \Vert_{d'} + \omega \times (1 - \Vert \mathcal{P}_{f} \Vert_{d'})}}_{\phi}
\end{equation*}

for some $\delta \geq \delta^{-} > 0$. Then defining the 4-tuple $(N, T, p, q)$ as

\begin{align*}
(p, q) &\in \mathbb{N}^{2} \textit{ such that } \phi \times (p + q) - p < 0\\
N &= -T \times (\phi \times (p+q) - p)\\
T &\in \mathbb{N} \textit{ such that } N \textit{ large enough}
\end{align*}

Then, reproducing the same development as it was done in the proof of previous theorem, one can derive a convenient form to easily apply the Chernoff inequality.

\begin{equation*}
\mathbb{P} \left( s_{v, f}(N, T, p, q) > 0 \vert \omega_{v, f} > \omega \right) \leq \mathbb{P}\left(\vert \sum_{t=0}^{T_i - 1} s_{v, p, q, t, f} - \mathbb{E}\left[s_{v, p, q, t, f}\right] \vert > T \times \delta_{f} \right)
\end{equation*}

With $\delta_{f} = (p + q) \times \delta^{-} \times \phi^{2} \times \frac{1 - \Vert \mathcal{P}_{f} \Vert_{d'}}{\Vert \mathcal{P}_{f} \Vert_{d'}}$. Then using the Chernoff inequality as written in \cite{TAO-1} using $\lambda = \sigma^{-1} \delta_{f}$ we obtain

\begin{align*}
\mathbb{P}\left(\vert \sum_{t=0}^{T - 1} s_{v, p, q, t, f} - \mathbb{E}\left[s_{v, p, q, t, f}\right] \vert > T \times \delta_{f} \right) &= \mathbb{P}\left(\vert \sum_{t=0}^{T - 1} s_{v, p, q, t, f} - \mathbb{E}\left[s_{v, p, q, t, f}\right] \vert > \lambda \sigma \sqrt{T} \right) \\
&\leq C \times \max \left( \exp\left(- T \times \left(\frac{\delta_{f} c}{\sigma}\right)^2 \right), \exp\left(- T \times \delta_{f}c \right) \right)
\end{align*}

With $C, c$ some positive constant and $\Var[s_{v, p, q, t, f}] = \sigma^2$. Q.E.D

\begin{center}
\rule[0pt]{100pt}{1pt} 
\end{center}

\subsection{Limit of the generic case}

The combination of theorems shows that the association of sampling and draining with the right choice of 5-tuple $(\omega, N, T, p, q)$ gives a convenient tool to select measure grid's bits with purity coefficient lower than a target $\omega$. Furthermore, when $T \rightarrow +\infty$, the correct selction is almost certain, which highlights the trade-off between efficiency and complexity of the algorithm that is embedded in the choice of $\omega$ and $T$, on which depends $N, p$ and $q$.
This generic procedure and its analysis deliver a strong framework that eases the derivation of more specific results that may be obtained under specific modelling of latent factors's activations and measure grid signatures. Nevertheless, it leaves two fundamental points clueless 

\begin{itemize}
\item No possibility to quantify further the effectiveness of the sampling strategy 
\item No specific procedure or heuristics to choose positive real value $\omega$
\end{itemize}

In the rest of this paper, we present two particular cases of factor's and measure grid's modelling that enables a better quantification of the sampling strategy and stronger heuristics for the choice of $\omega$. 

\section{Case of signal plus noise}

This particular case is designed to be easy to analyze. We first define the statistical modelling of factors and bits's activations. Then, we quantify the sampling strategy and justify a choice for the 5-tuple $(\omega, T, N, p, q)$. Finally, we present simulations and provide discussion of the results obtained with this special case.

\subsection{Statistical modelling}

In this particular case, we assume that the target factor $f$ is linked to some $\vert \mathcal{G}(f)\vert = k$ measure grid's bits and activates with probability $p_f$. We also assume that bits of the measure grid are identically and independently subject to a noisy activation with probability $p_N$. We may see noisy activations as the result of $n$ noisy latent factors, linked to exactly 1 bit of the measure grid, that is $K=n + 1$. Under this model, the probability for a bit $b\in \mathcal{G}$ to activate is defined as 

\begin{equation*}
\mathbb{P} \left( \textit{"b active"} \right) = \begin{cases} p_{f} + p_N \times (1-p_{f}), & \text{if }\ b \in \mathcal{G}(f^{*}) \\ p_N, & \text{Otherwise} \end{cases}
\end{equation*}

As a consequence, for any $I \in S(\mathcal{G})$ such that $\vert I \cap \mathcal{G}(f) \vert = i$ and $j = \vert I \vert -i$ if we set $x \in F_2 (\{I\} , \mathcal{G})$, the distribution over measure grid bits's activations is defined as

\begin{equation*}
d_{x} =  \begin{cases} p_N^{i+j} \times (1-p_N)^{n - i - j} \times (1- p_{f}) + p_N^{j} \times (1-p_N)^{n - k - j} \times p_{f}, & \text{if }\ i=k \\ p_N^{i+j} \times (1-p_N)^{n - i - j} \times (1- p_{f}), & \text{Otherwise} \end{cases}
\end{equation*}

In the rest of the section, we will always refer to this distribution as $d$.

\subsection{Evaluation of bits}

Let $G$ be a firing graph with a layer 0 composed of measure grid's bits. Then, the precision of a vertex $v(I,  \vert I \vert)$ of layer 1 of $G$, with respect to $f$, depends only on $\vert I \vert$ and $\vert I \cap \mathcal{G}(f) \vert$. Indeed, if $\vert I \cap \mathcal{G}(f) \vert = i$ then

\begin{equation*}
\phi_{v,f} = \frac{p_{f}}{p_{f} + (1 - p_{f})\times p_N^{i}}
\end{equation*}

With identification of terms using (\ref{prop:precision1}) we have $\omega_{v, f} = p_N^{i}$ and using previously defined distribution, one finds that $\mu_{v, f} = p_N^{\vert I \vert - i}$, $\nu_{v, f} = p_N^{\vert I \vert}$. Besides, given a set of bits $I$ such that  $I \subset \mathcal{G}(f)$, if $b \in \mathcal{G}(f) \setminus I$, the precision of vertex $v(I \cup \{ b \}, \vert I \vert +1)$ with respect to $f$ is 

\begin{equation*}
\phi_{v} = \frac{p_{f}}{p_{f} + (1 - p_{f})\times p_N^{\vert I \vert + 1}}
\end{equation*}

if  $b \notin \mathcal{G}(f)$

\begin{equation*}
\phi_{v} = \frac{p_{f}}{p_{f} + (1 - p_{f})\times p_N^{\vert I \vert}}
\end{equation*}

\subsection{Sampling Strategy}

In this particular case we follow the generic sampling procedure $\mathcal{S}$ with parameter $p_{\mathcal{S}}$. Thus, using the previously defined statistical distribution of bits's activations, if we denote $S$, the set of sampled bits using $\mathcal{S}$, the distribution of the cardinal of $S$ is 

\begin{align*}
\mathbb{P}\left(\vert S \vert = s\right) = \begin{cases} \begin{pmatrix} n - k \\ s - k  \end{pmatrix} \times p_N^{s - k } \times (1 - p_N)^{n-k-s} \times p_{\mathcal{S}} , & \text{if }\ s \geq k \\ 0, & \text{otherwise} \end{cases}
\end{align*}

Thus its expected size is $\mathbb{E}\left[ \vert S \vert \right] = k + (n - k) \times p_N \times p_{\mathcal{S}} $. Furthermore if $I= \{b_1^{*}, \ldots, b_i^{*} \}\in S(\mathcal{G})$ is some set of pre-selected bits and $S$ is a set of bits sampled using $\mathcal{S}$, a positive real $\omega_i$ exists such that

\begin{align*}
\mathbb{P}_{\mathcal{S}} \left( \vert \lbrace s \in S \setminus \omega_{I \cup \{s\}, \vert I \vert + 1, f } < \omega_i \rbrace \vert = j \right)  &= \mathbb{P}_{\mathcal{S}} \left( \vert \lbrace s \in S \setminus s\in \mathcal{G}(f) \vert = j \right)\\
&= \begin{pmatrix} \vert \mathcal{G}(f) \vert -i \\ j  \end{pmatrix} \times p_{\mathcal{S}}^{j} \times (1 - p_{\mathcal{S}})^{\vert \mathcal{G}(f) \vert -i - j}
\end{align*}

\subsection{Identification of factors}

First, in the case of a single sampled firing graph, one can see that bits's purity coefficients take only two values with respect to $f$

\begin{align*}
\omega_{\{b\}, 1, f} = \begin{cases} p_N , & \text{if }\ b \in \mathcal{G}(f) \\ 1, & \text{otherwise} \end{cases}
\end{align*}

Thus if we choose 

\begin{equation*}
\omega_0 = \frac{(1 + p_N)}{2}
\end{equation*}

It maximizes the purity margin defined as
 
\begin{align*}
\delta_0 &=  \frac{(\omega_0 - \omega_{\{ b \}, 1, f}) + (\omega_{\{ b' \}, 1, f} - \omega_0)}{2} = \frac{(1-p_N)}{2}
\end{align*}

Where $b \in \mathcal{G}(f)$ and $b' \notin \mathcal{G}(f)$. In the case of a joint sampled firing graph in which a set $I = \{b_1^{*}, \ldots,  b_i^{*}\}$ of $i \in \mathbb{N}^{*}$ pre-selected bits that verify $\forall b \in I$, $b \in \mathcal{G}(f)$, remaining bit's purity coefficients with respect to $f$ can take again two values

\begin{align*}
\omega_{I \cup \{b\}, i+1, f} = \begin{cases} p_N^{i+1} , & \text{if }\ b \in \mathcal{G}(f) \\ p_N^{i}, & \text{otherwise} \end{cases}
\end{align*}

Thus if we choose

\begin{equation*}
\omega_i = \frac{(1 + p_N) \times p_N^{i}}{2}
\end{equation*}

it maximizes the purity margin defined as
 
\begin{align*}
\delta_i &=  \frac{(\omega_i - \omega_{I \cup \{ b \}, i+1, f}) + (\omega_{I \cup \{ b' \}, i+1, f} - \omega_i)}{2} = \frac{(1-p_N) \times p_N^{i}}{2}
\end{align*}

Where $b \in \mathcal{G}(f)$ and $b' \notin \mathcal{G}(f)$. Finally we define the 5-tuple $(N_i, T_i, p, q)$ as 

\begin{align*}
(p, q) &\in \mathbb{N}^{2} \textit{ such that } \phi_i \times (p + q) - p \leq 0 \textit{ and } \phi_i' \times (p + q) - p > 0 \\
N &= -T \times (\phi \times (p+q) - p)\\
T &\in \mathbb{N} \textit{ such that } N \textit{ large enough}
\end{align*}

Where $t \in \mathbb{N}$, $\phi_i = \frac{p_{f}}{p_{f} + \omega_i \times (1 - p_f)}$ and $\phi_i' = \frac{p_{f}}{p_{f} + (\omega_i - \delta_i) \times (1 - p_f)}$.

\subsection{Simulation}

The signal plus noise model is implemented in python and mainly uses standard numpy and scipy modules to generate random signal that fit its probabilistic model. More details about the implementation can be found in appendix B. We generate $n=1000$ bits that randomly activate with probability $p_N$ and we choose randomly $\vert  \mathcal{G}(f)\vert = 50$ bits that are linked to a latent factor that activates with probability $p_f=0.3$. Finally we build the single sampled firing graph using $p_{\mathcal{S}} = 1$.

\begin{figure}[H]
\subfloat[$p_N = 0.3$, $(p,q)=(1,1)$]{\includegraphics[scale=0.165]{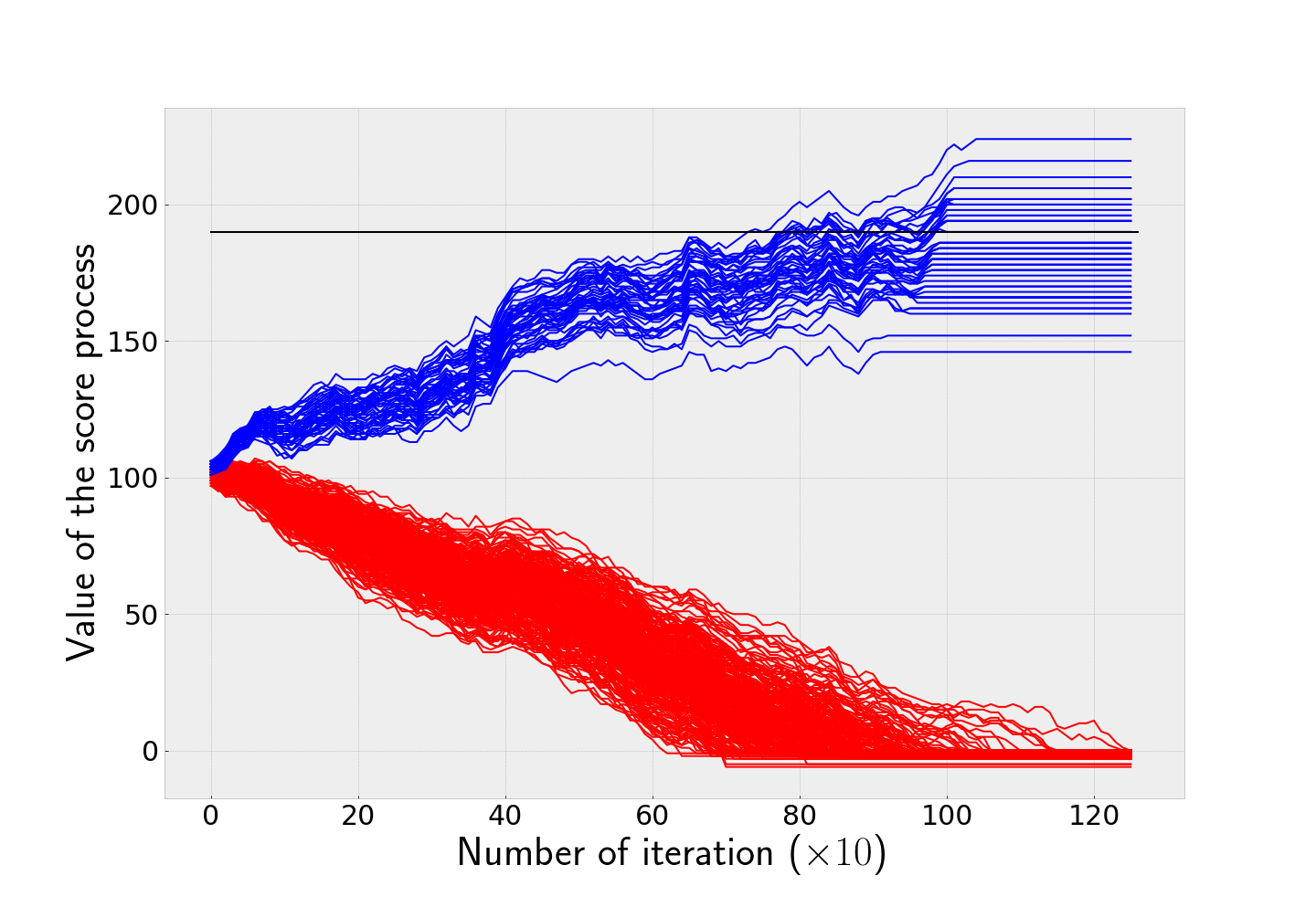}} 
\subfloat[$p_N = 0.5$, $(p,q)=(2,3)$]{\includegraphics[scale=0.165]{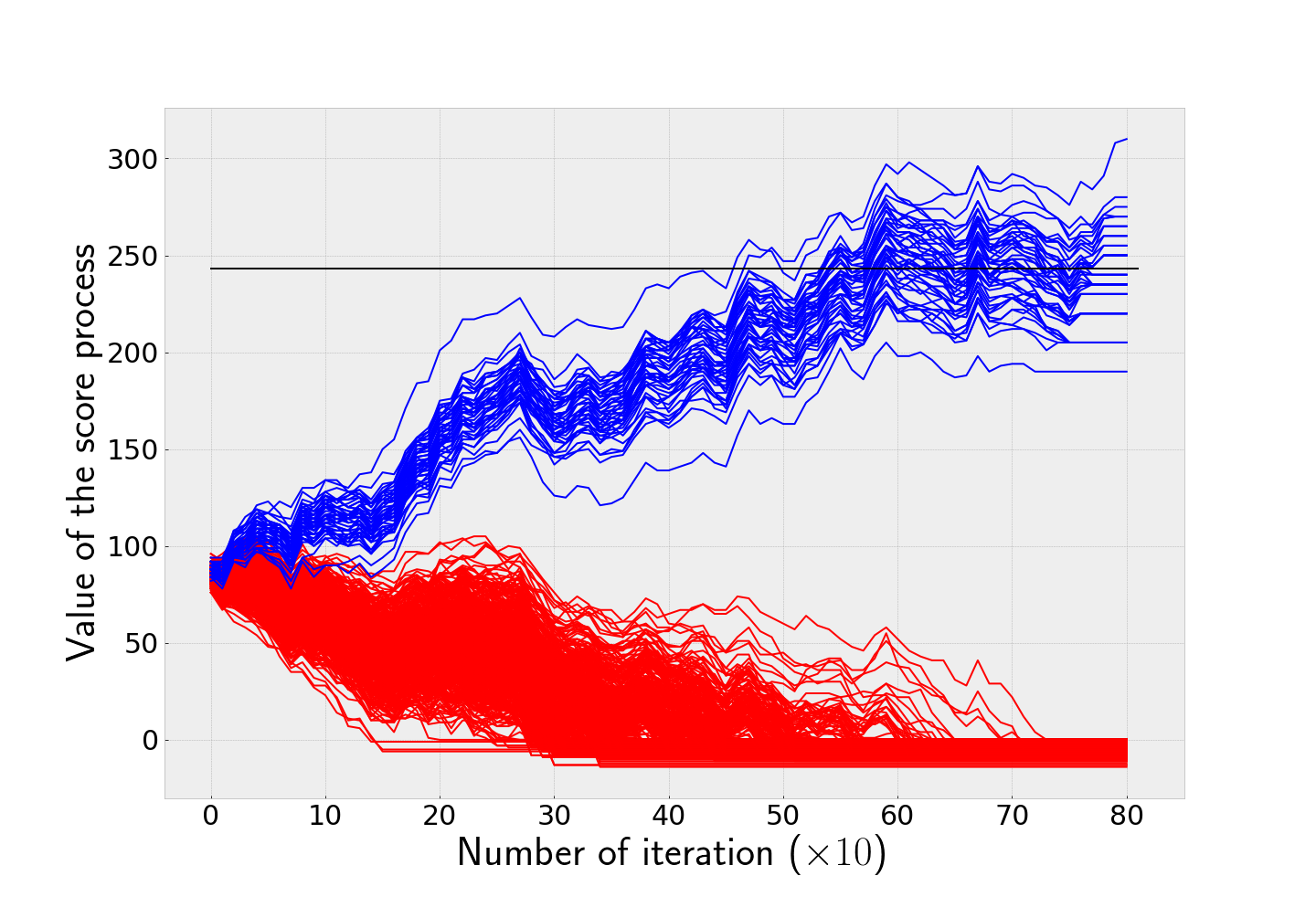}}\\
\subfloat[$p_N = 0.7$, $(p,q)=(3, 5)$]{\includegraphics[scale=0.165]{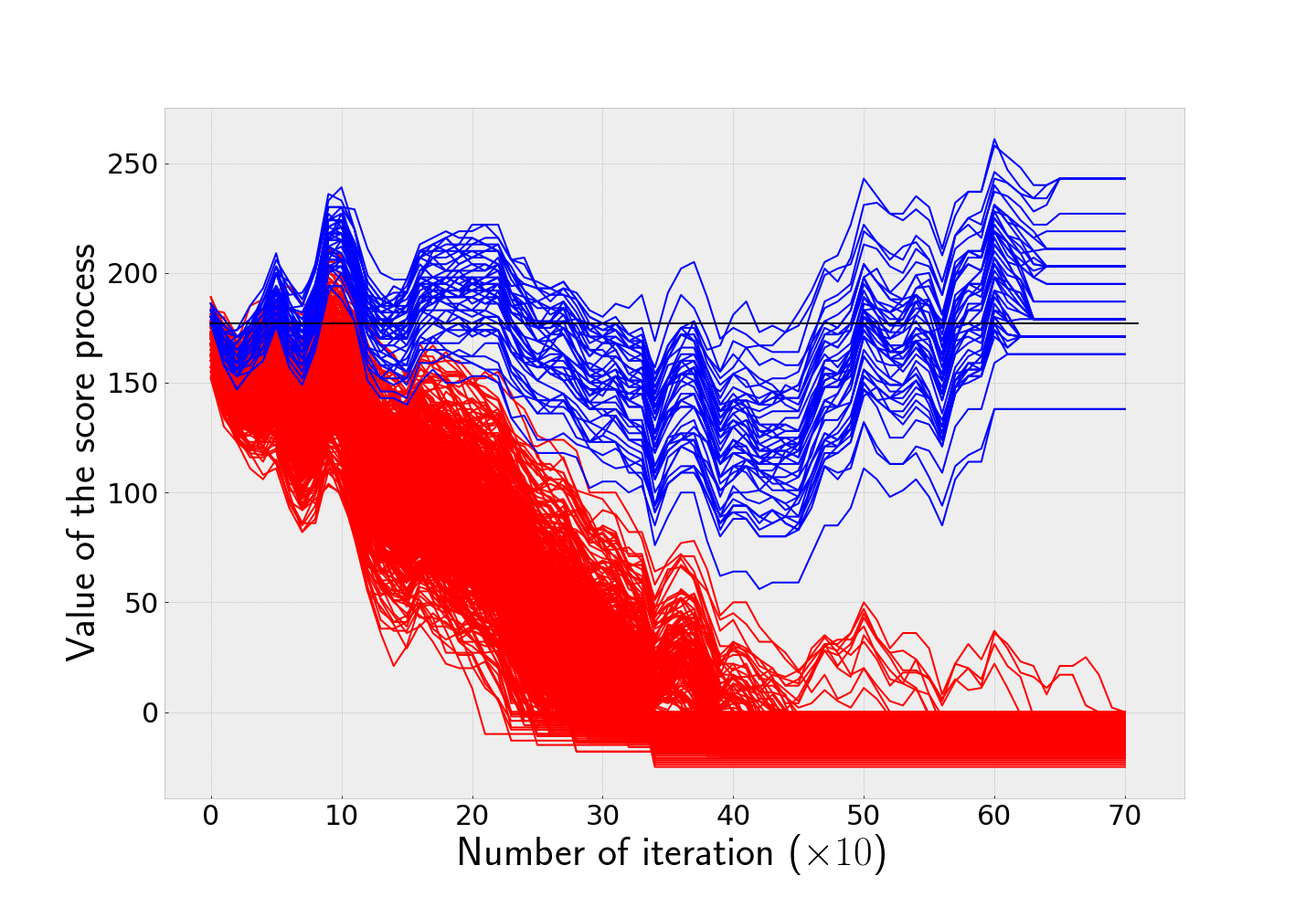}}
\subfloat[$p_N = 0.9$, $(p,q)=(5,11)$]{\includegraphics[scale=0.165]{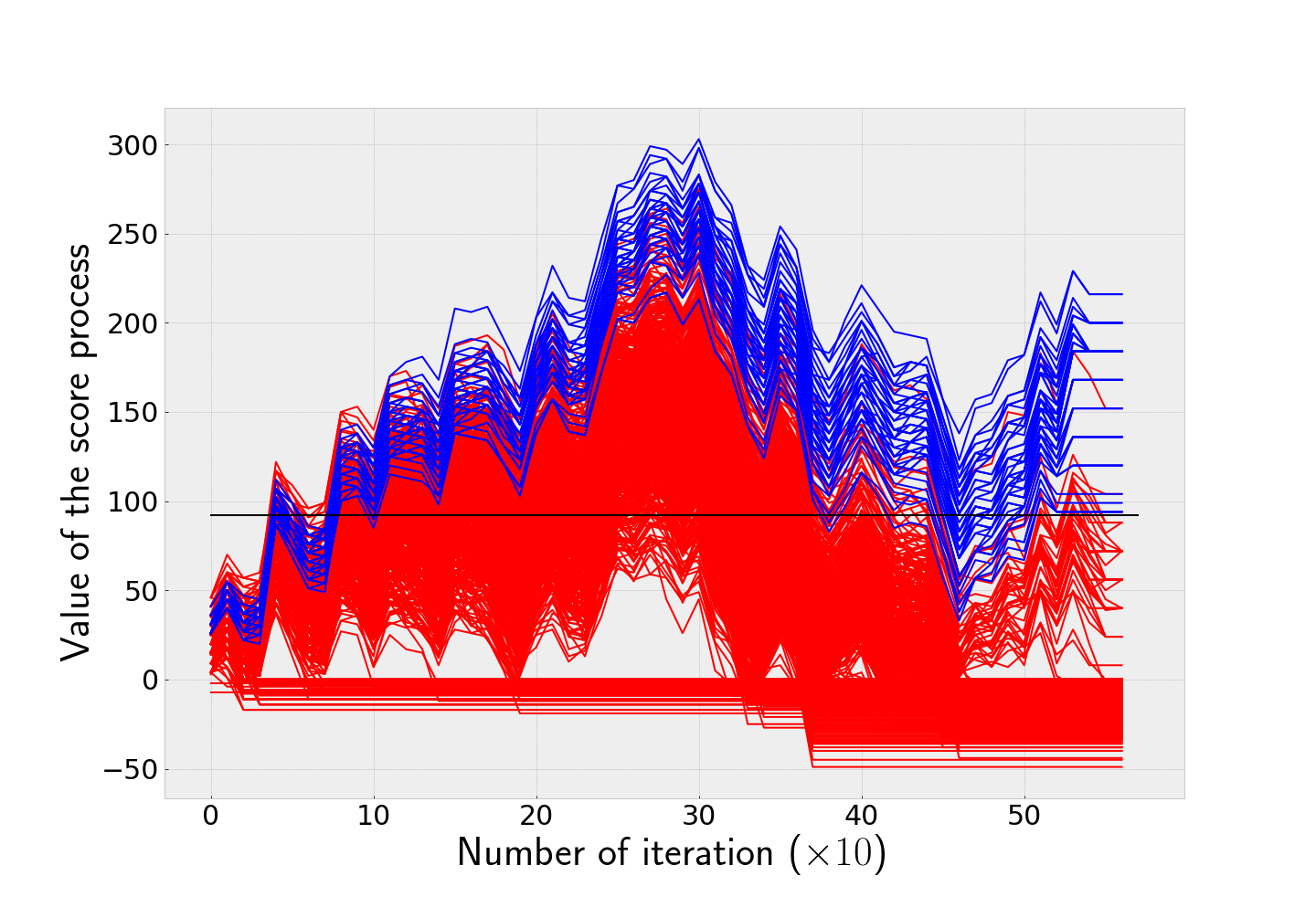}} 
\caption{Observation of the score process for different SNR models $T=500$}
\label{fig:sim_spn_1}
\end{figure}

Each subplot of figure~\ref{fig:sim_spn_1} shows the weight of outcoming edges of sampled vertices. Blue lines show the weight of edges outcoming from sampled bit $b \in \mathcal{G}(f)$ and red lines correspond to the weight of edges outcoming from sampled bit $b \notin \mathcal{G}(f)$. Finally the black horizontal line represents the theoretical mean value of $s_{v, f}[N, T, p, q]$ of a vertex with characteristic polynome $\mathcal{P}_v = \mathcal{P}_{\{b\}, 1}$, with $b \in \mathcal{G}(f)$. As theory suggests, we can see two distinct phenomenons, blues lines converge around theoretical mean for process of bits linked to the target factor and red lines converge to 0. However, the higher is $p_N$,  the less noticeable is the distinction between each process. This is explained by the fact that the higher is $p_N$, the closer are the precision of bits linked to target factor $f$ and the precision of noisy bits. Futhermore the later observation induces a high value of $p+q$ which result in a more volatile score process. For the second simulation we use $n=1000$, $\vert \mathcal{G}(f)\vert = 50$, $p_f=0.3$, $T=200$ and $p_{\mathcal{S}}= 0.5$. Yet at the end of the draining we choose all the input vertices of the firing graph that still have an outcoming edge and use their combined activation as an estimator of the target factor's activation. We then measure their precision and recall over $100$ repetion for each SNR ratio

\begin{table}[H]
\begin{tabular}{|l|l|l|l|l|l|}
\hline
$P_N$ & Mean $\phi$ & Standard deviation $\phi$ & Mean $\psi$ & Standard deviation $\psi$ & Number of fails \\ \hline
0.3   & 1.0         & 0.00                       & 0.87         & 0.30                      & 0               \\ \hline
0.5   & 1.0         & 0.03                      & 0.66         & 0.42                      & 0               \\ \hline
0.7   & 0.97        & 0.13                      & 0.43        & 0.46                      & 4               \\ \hline
0.9   & 0.75        & 0.14                      & 0.13        & 0.25                      & 19              \\ \hline
\end{tabular}
\caption{Evaluation of naive factor's activation estimation, $T=200$, $100$ repetitions}
\label{tab:sim_spn_1}
\end{table}

Table~\ref{tab:sim_spn_1} shows quality indicators of the estimator for different SNR ratio. The two first columns give respectively the mean and standard deviation of the precision of the estimator. The two following columns are respectively the mean and the standard deviation of the recall of the estimator. Finally the last column is the number of experiments that ended without any input vertices having a path towards the output, so that the construction of an estimator is not possible. Again, we see that the quality of the estimator drops as the theoretical precision between noisy bits and factor's bits are close to each other. Yet it reveals that this naive estimator, for a reasonable SNR ratio, is still efficient to predict the activation of target latent factor. Finally, we simulate the signal plus noise model in the settings of joint sampled firing graph. We use a measure grid of $n=1000$ bits from which we sampled randomly $\vert \mathcal{G}(f) \vert = 50$ bits linked to target factor $f$ that activates with probability $p_f = 0.3$ and we set $p_N=0.6$. Finally we built the joint sampled firing graph by pre-selecting randomly 5 bits linked to the factor and running the sampling algorithm described previously using $p_{\mathcal{S}} = 1$.

\begin{figure}[H]
\centering
\includegraphics[scale=0.30]{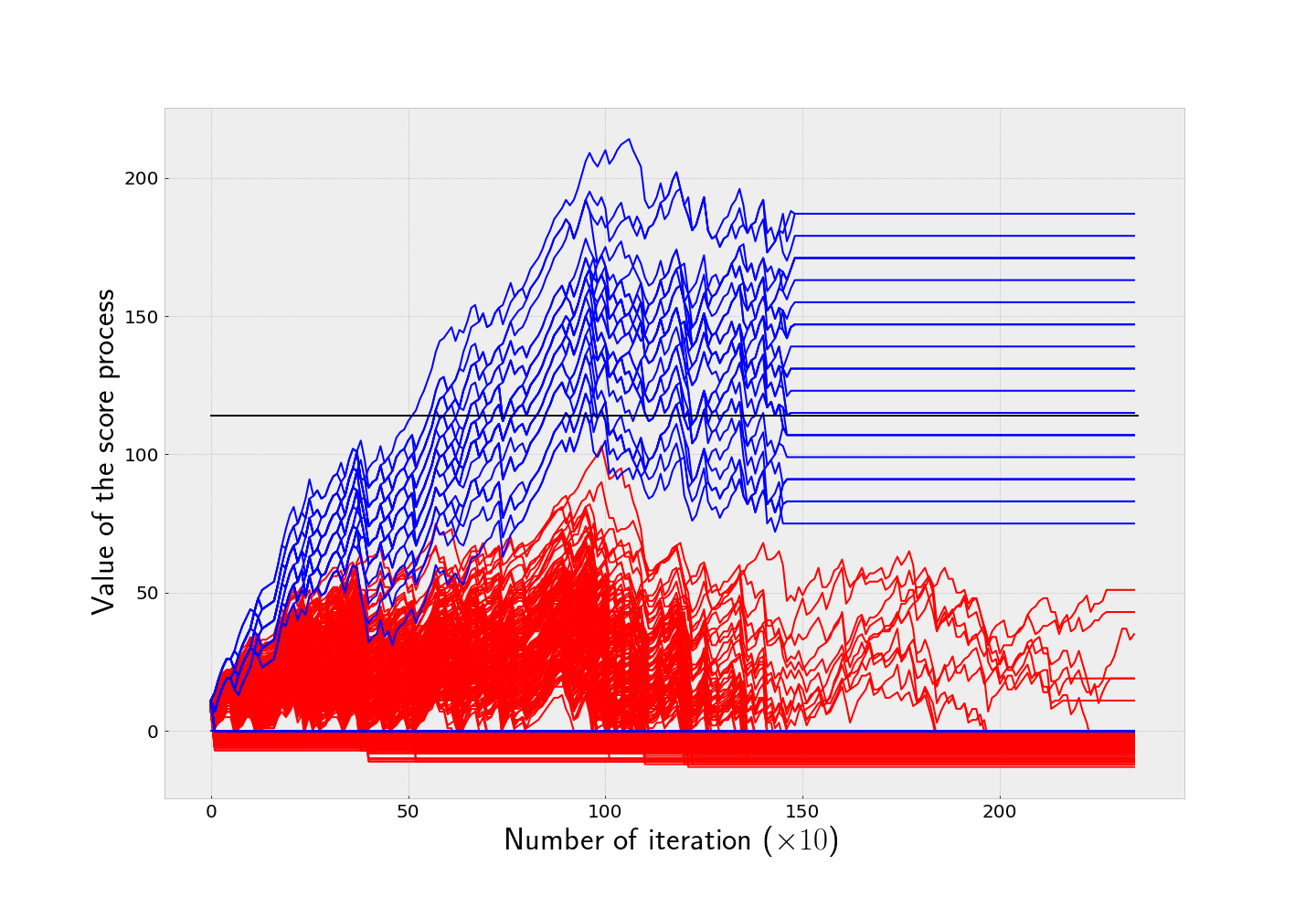}
\caption{Observation of the score process in a joint sampled firing graph with $T=500$ and $(p, q) = (7, 1)$}
\label{fig:sim_spn_2}
\end{figure}

In this case, we obtain $N=7$ and $\omega_5 \simeq 0.062$ when following the procedure described in previous section. As for the first experiment, blue lines show the weight of edges outcoming from sampled bit $b \in \mathcal{G}(f)$ and red lines correspond to the weight of edges outcoming from sampled bit $b \notin \mathcal{G}(f)$. The black horizontal line represents the theoretical mean value of $s_{v, f}[N, T, p, q]$, where $v$ has characteristic polynome $\mathcal{P}_v = \mathcal{P}_{\{b_1, \ldots, b_5, b\}, 5}$, with $\{b_1, \ldots, b_5\}$ the set of pre-selected bits and $b \in \mathcal{G}(f)$. The simulation validate the expectation from theory and the high value of $p+q$ explains the high volatility of score processus.

\section{Case of sparse measure grid}

This particular case is more complex than the previous one. We first define the statistical signature of factors and bits's activations. Then we quantify the sampling strategy and justify a choice for the 5-tuple $(\omega, T, N, p, q)$. Finally, we present simulations and provide discussion of results obtained with this particular case.

\subsection{Statistical modelling}

\subsubsection*{Latent factor activation}

We assume that each of the $K$ latent factors activates independantly with probability $p_f$. As a consequence, for any $I \in S(\mathcal{F})$, if we define $x$ such that $x \in F_2 (\{I\}, \mathcal{F})$, we can define the distribution of factor's activation as 

\begin{equation*}
d'_x = \begin{pmatrix} \vert I \vert \\ K \end{pmatrix} \times p_f^{\vert I \vert} \times (1-p_f)^{K-\vert I \vert} 
\end{equation*}

\subsubsection*{Measure grid activation}

We assume two major properties of activations of measure grid's bits.

\begin{itemize}
\item For each factor $f \in \mathcal{F}$, each bit $b \in \mathcal{G}$ has equal probability $p_g$ to belong to $\mathcal{G}(f)$.
\item For each factor $f \in \mathcal{F}$, for each couple $b_1, b_2 \in \mathcal{G}^2$, events "$b_1 \in \mathcal{G}(f)$" and "$b_2 \in \mathcal{G}(f)$"  are independent.
\end{itemize}   

As a consequence the probability for a bit $b$ to activate, given that every factor of some set $\{f_1, \ldots, f_k\} \subset \mathcal{F}$ is active, writes

\begin{align*}
\mathbb{P}\left(\textit{b active } \vert f_1, \ldots, f_k \textit{ active}\right) &= \sum_{i=1}^{k} \begin{pmatrix} k \\ i \end{pmatrix} \times p^i_g \times (1 - p_g)^{k - i}\\
 &= 1 - (1 - p_g)^{ k}
\end{align*}

The above quantity depends only on the number of active latent factors. Thus, for any $I \in S(\mathcal{G})$, if we define $x \in F_2 (S(\mathcal{G}) , \mathcal{G})$, we can define the distribution of bits's activations as 

\begin{equation*}
d_x = \sum_{k = 1}^{K} \left[ \begin{pmatrix} K \\ k \end{pmatrix} \times p_f^{k} \times (1-p_f)^{K-k}\right] \times \left[ p^i_{g \vert k} \times (1 - p_{g\vert k})^{n - i} \right]
\end{equation*}

With $i = \vert I \vert$ and $p_{g \vert k} = \mathbb{P}\left(\textit{b active } \vert f_1, \ldots, f_k \textit{ active}\right)$.

\subsection{Evaluation of bits}

Let $G$ be a firing graph whose layer 0 is composed of measure grid's bits. Given a target factor $f$, the precision with respect to $f$ of a vertex $v(\{ b \}, 1)$ of the layer 1 of $G$ depends on wether $b \in \mathcal{G}(f)$ and on $\vert \mathcal{G}^{-1}(b) \vert$. Indeed, if $b \in \mathcal{G}(f)$ and $\vert \mathcal{G}^{-1}(b) \vert = l $, $l \in \{1, \ldots, K\}$, it will be said to have a purity rank of $l$ and its precision with respect to $f$ writes

\begin{equation*}
\phi_{v, f} = \frac{p_f}{p_f + (1 - p_f)\times \omega_{l}}
\end{equation*}

where $\omega_l = 1 - (1-p_f)^{l-1}$. If $b' \notin \mathcal{G}(f)$ the precision of $v'(\{ b' \}, 1)$ with respect to $f$ writes

\begin{equation*}
\phi_{v', f} = p_f
\end{equation*}

Futhermore, if we have a vertex $v(I, \vert I \vert)$ such that $\forall b \in I$, $b \in \mathcal{G}(f)$ and $\min_{b \in I} \vert \mathcal{G}^{-1}(b) \vert = l$, then the precision of $v$ with respect to $f$ verifies

\begin{equation*}
\phi_{v, f} \leq \frac{p_f}{p_f + \omega^{-}_l \times (1 - pf)}
\end{equation*}

With 

\begin{equation*}
\omega^{-}_l = \sum_{k=K-l-1}^{K} \begin{pmatrix} K \\ k \end{pmatrix} p_f^{k} \times (1-p_f)^{K - k}
\end{equation*}

The minimum purity coefficient one can obtained with bits that verifies $b \in \mathcal{G}(f)$ and $\vert \mathcal{G}^{-1}(b) \vert = l$. That is, the case of a vertex $v(I, \vert I \vert)$ with $I$ composed of every possible $\begin{pmatrix} K \\ l \end{pmatrix}$ such bits.

\subsection{Sampling Strategy}

We follow the generic sampling procedure $\mathcal{S}$ with parameter $p_{\mathcal{S}}$. Although it is not hard to derive key quantification such as $\mathbb{E}\left[ \vert S \vert \right]$ or probabilities to sample bits linked to a target factor $f$ under this modelling, generic formulas are not elegant and present not much interest in this simulation.
 
\subsection{Identification of factors}

First, in the case of a single sampled firing graph, for any grid's bits linked to factor $f$, there is only $K$ different purity coefficients possible. Thus we may set $\omega$ to $\omega_l$, using $l$  reasonably small to differentiate lower purity rank from greater purity rank samples. In the case of a joint sampled graph, where a set of $I=\{b_1^{*}, \ldots b_i^{*}\}$ were pre-selected then the choice of $\omega$ is not trivial and is hard to be efficiently and generically derived. Let $\omega_{I, \vert I \vert, f}$ the purity coefficient of the pre-selected set of bits we set $\omega = \omega_{I, \vert I \vert, f} - \delta$ where $\delta \in \mathbb{R}$ should be chosen with caution. Finally we choose the 5-tuple $\lbrace (\omega, N, T, p, q)$ as 

\begin{align*}
(p, q) &\in \mathbb{N}^{2} \textit{ such that } \phi \times (p + q) - p < 0 \\
N &= -T \times (\phi \times (p+q) - p)\\
T &\in \mathbb{N} \textit{ such that } N \textit{ large enough}
\end{align*}

Where $\phi = \frac{p_f}{p_f + \omega \times (1-p_f)}$. 

\subsection{Simulation}

The sparse measure grid model is implemented using python and the standard python numpy and scipy modules to generate random signal that fit its probabilistic model. In our case we generate $n=1000$ bits with $K=10$ latent factors that activate with probability $p_f = 0.3$ and we link measure grid's bits independently with probability $p_g=0.3$. Finally we built the single sampled firing graph running the sampling algorithm described previously, using $p_{\mathcal{S}} = 1$. Finally, we set $\omega= \omega_{10}$, the higher purity coefficient for bits linked to the target factor $f$.

\begin{figure}[H]
\centering
\includegraphics[scale=0.33]{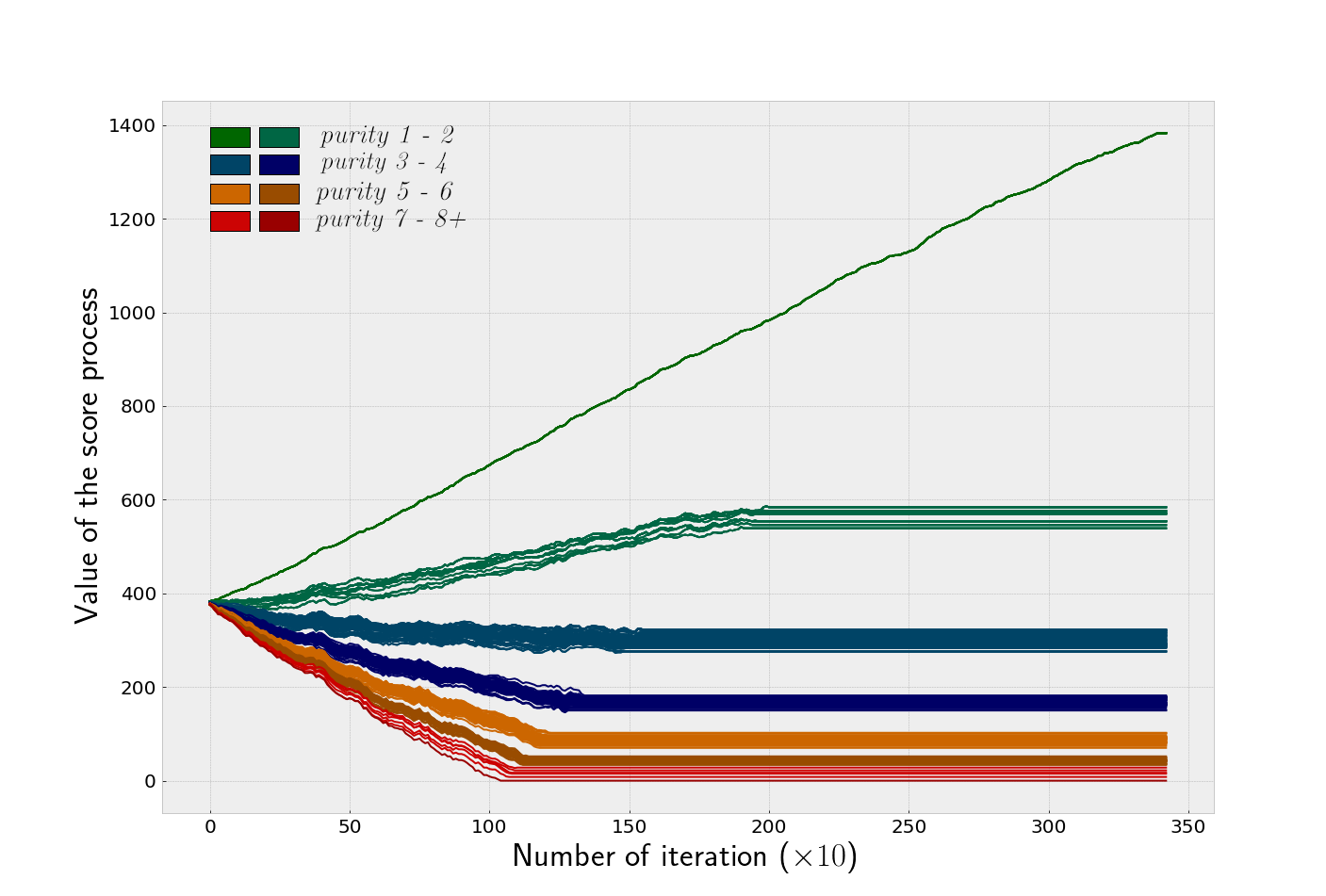}
\caption{Observation of the score process in a single sampled firing graph with $T=1000$ and $(p, q) = (1, 1)$}
\label{fig:sim_si_1}
\end{figure}

We clearly see a rapid differentiation of score processus according to their purity rank. We can also observe that, at the end of draining, the higher the purity coefficient is, the closer are weights of corresponding edges. Finally, the behaviour of score processus validates the efficiency of the draining algorithm to rank bits of the measure grids blindly, in an attempt to identify latent factors. The second experiment with the sparse measure grid model aims to give intuition on the choice of $\delta$ used for draining a joint sampled firing graph. As for the previous simulation, we generate $n=1000$, bits with $K=10$ latent factors that activate with probability $p_f = 0.3$ and we link measure grid's bits independently with probability $p_g=0.3$. Then we choose randomly $i=5$ bits, denoted by $I=\{b_1^{*}, \ldots, b_i^{*}\}$, with purity rank $4$ with respect to the target factor $f$. Finally we sampled and built the joint sampled firing graph using $p_{\mathcal{S}} = 1$ and the set of pre-selected bits $I$. The procedure described in the previous section to choose the target purity coefficient consists in estimating the purity coefficient $\hat{\omega}_{I, \vert I \vert, f}$ and to set $\delta$ so that $\omega = \hat{\omega}_{I, \vert I \vert, f} - \delta$.\\

\begin{figure}[H]
\subfloat[$\delta = 0.$]{\includegraphics[scale=0.165]{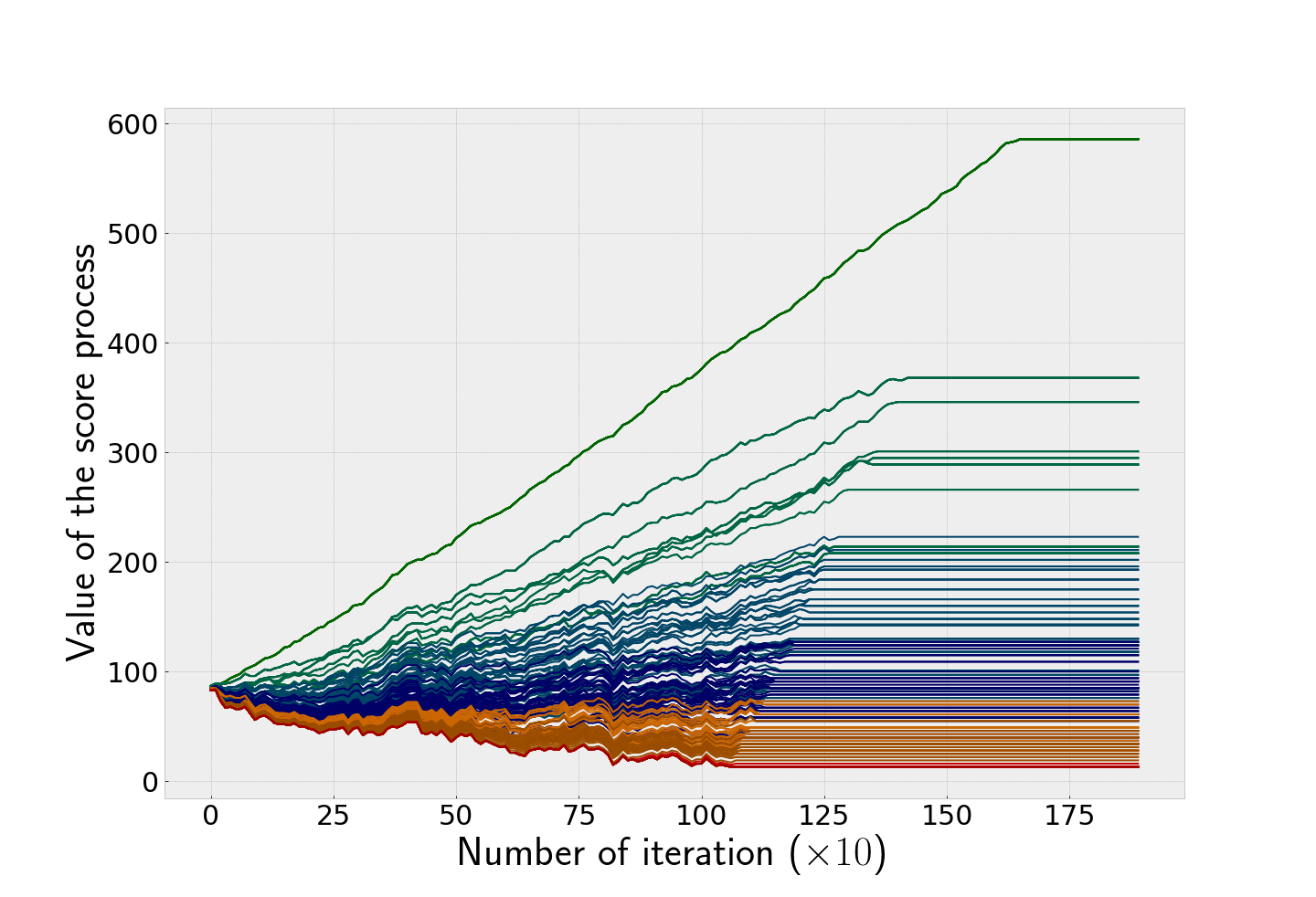}} 
\subfloat[$\delta = 10^{-2}$]{\includegraphics[scale=0.165]{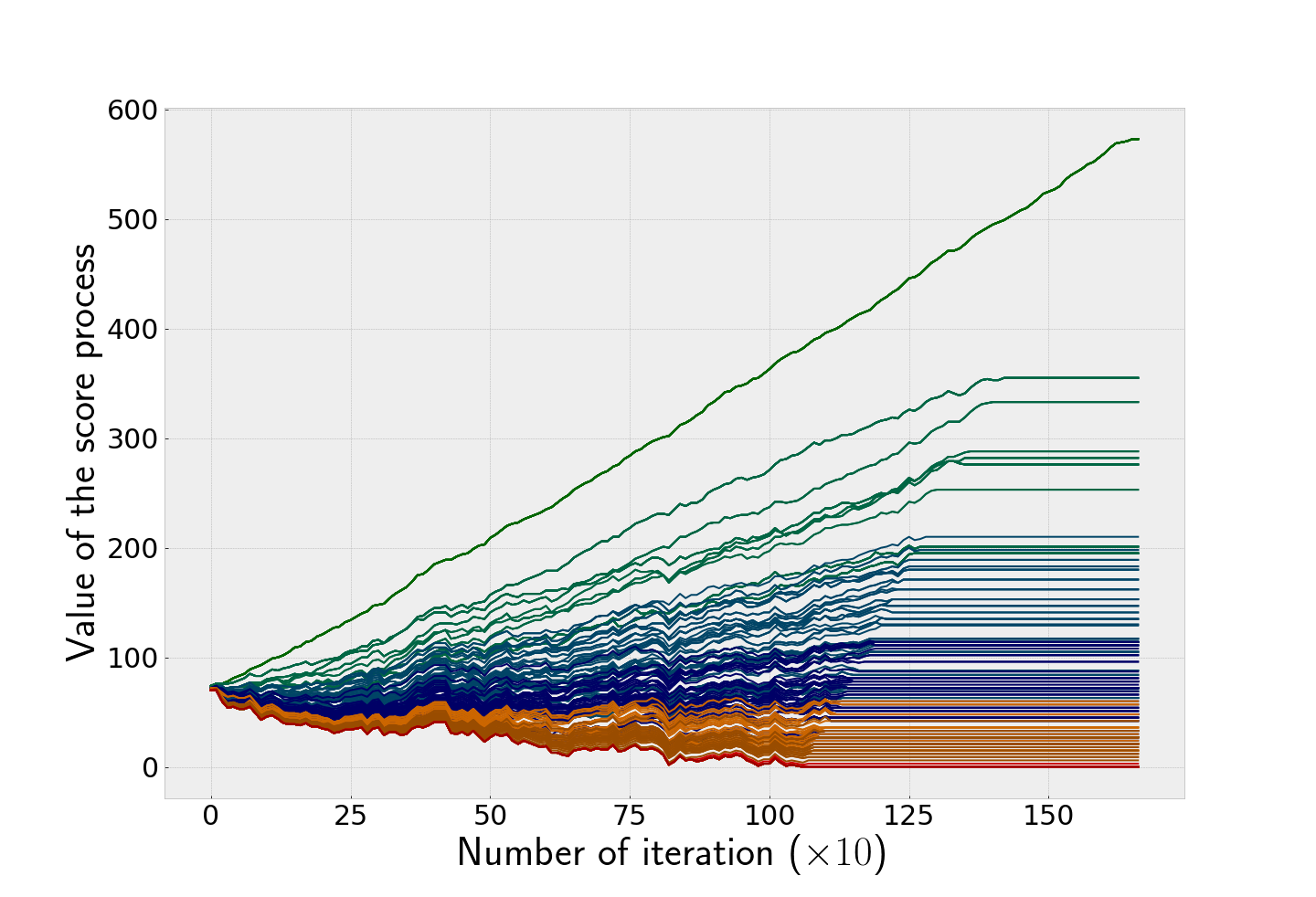}}\\
\subfloat[$\delta = 5 \times 10^{-2}$]{\includegraphics[scale=0.165]{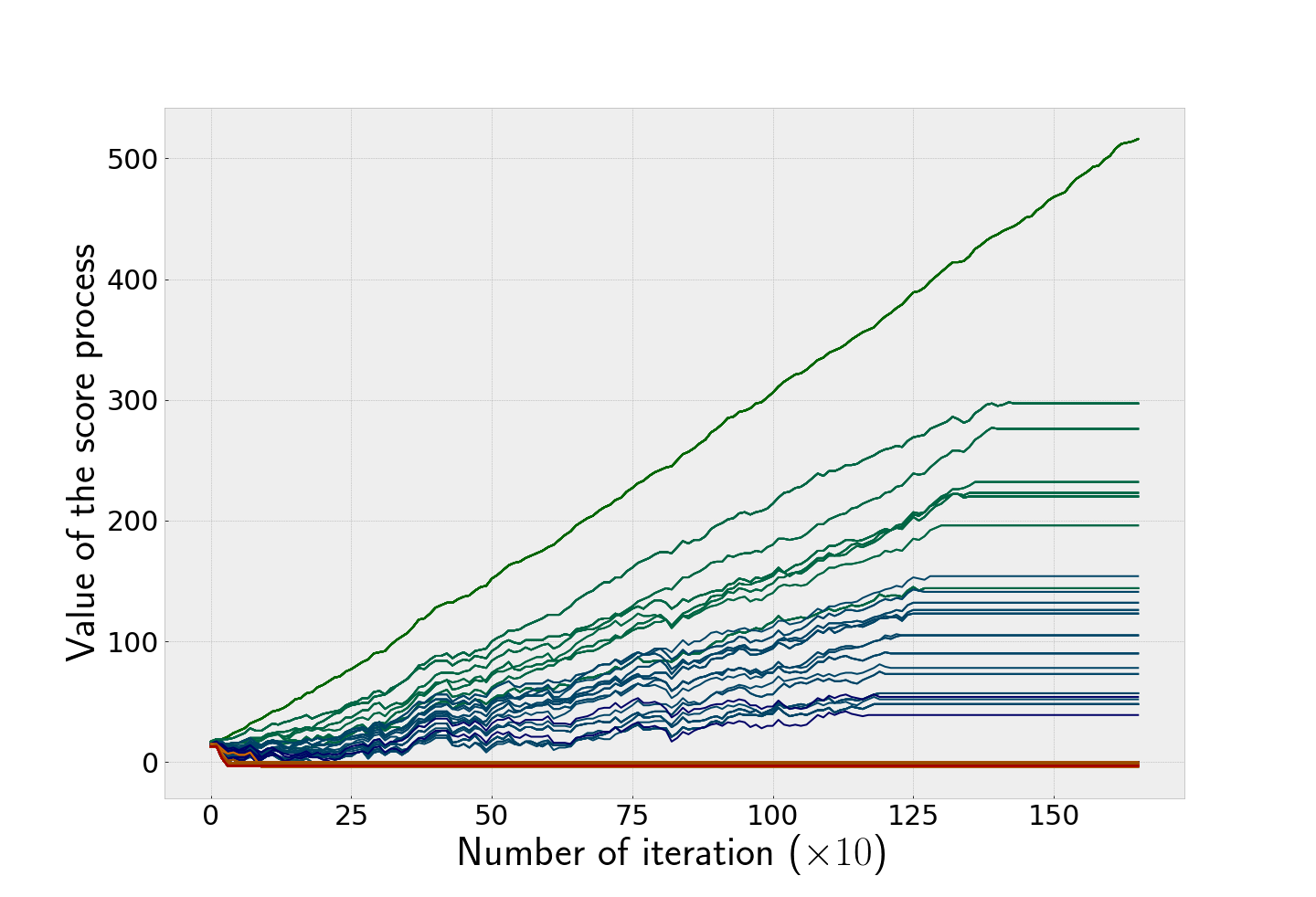}}
\subfloat[$\delta = 10^{-1}$]{\includegraphics[scale=0.165]{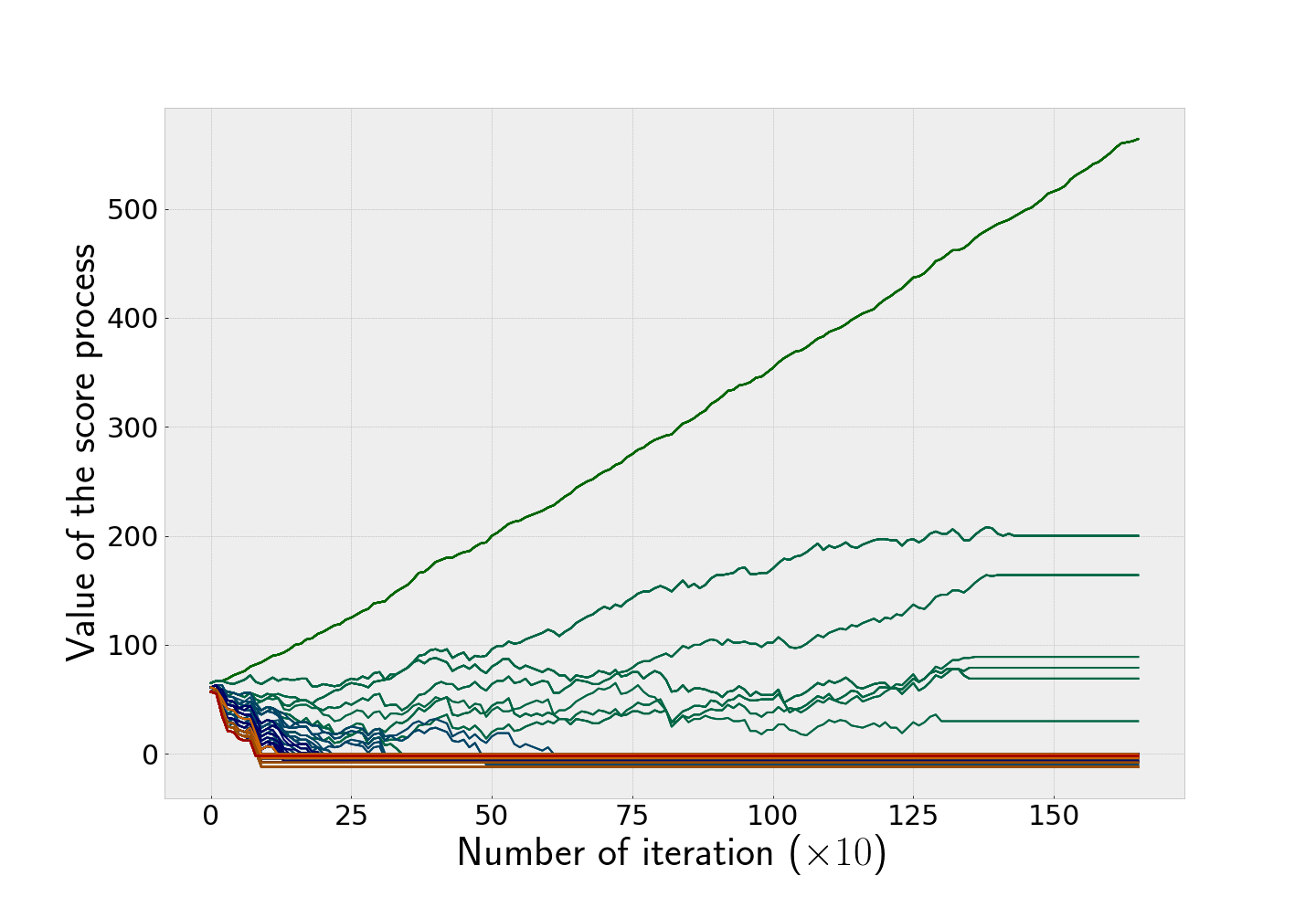}} 
\caption{Observation of sampled bit's score processus in a joint sampled firing graph $i=5$, $T=500$ and $(p, q) = (1, 1)$}
\label{fig:sim_si_2}
\end{figure}

In each simulation, $\hat{\omega}_{I, \vert I \vert, f}$ has been estimated using $1000$ samples and we use $T=500$. Furthermore, each figure corresponds to a different value of $\delta$ that induces different values of $\omega$, set as $\omega = \hat{\omega}_{I, \vert I \vert, f} - \delta$. As for the first experiment, the different colored lines in each subfigure show the weight of edges outcoming from sampled bits with different purity ranks. As expected, we can see that the higher $\delta$ is, the more discriminative the draining procedure is.  If $\delta$ is set to 0, then every sampled bits will remain connected in the firing graph after draining, which is not of great interest. Yet, if $\delta$ is set too high we may end with two connexe components, which is not desirable neither. Thus, the experiment confirms the difficulties that we may face choosing the right value for $\delta$.

\section{Discussion}

This paper has presented an algorithm that consists in a generic optimisation of a firing graph, in an attempt to solve the abstract task of identifying latent factor's activations. Furthermore it has provided theoretical certitude on the effectivness of the procedure. However, the iterative optimisation method associated with the diversity and flexibility of the architecture of a firing graph opens doors to further applications, notably in the field of inverse problem and in the very hype field of machine learning. Indeed in supervised classification, we are given a dataset composed of features that may be numerical or categorical description of samples and targets that specify the class of samples. If we assume that the activation of a target is a combination of latent factors's activations and that we operate the minimum transformation of features so that they take the form of a measure grid, a light layer of procedures could turn our solution into a supervised classificator. The specificity of such a learner would give it an interesting position in the supervised learning landscape. Indeed, its iterative optimisation and flexible architecture could make it an adaptative learner, that scale to large dataset, with minimum processing work on raw data, in the manner of a neural network. Yet unlike neural network the algorithm handle very efficiently categorical or sparse feature space. Furthermore, compared to the most advanced tree based classification, its flexible architecture is more suitable to learning update and on-the-fly evaluation or addition of new features. Finally, given the hype granted to the field of machine learning nowaday, both in the scientific comunity and civil society, it would be common sense to orient this piece of research to this field.

\newpage
\begin{center}
\LARGE \textbf{Appendices}
\end{center}
\appendix

\section{Properties}

\underline{Partition}\\

Let $v_1(I, l_0)$, $v_2(J, 0)$ and $v_3(K, 0)$, be three vertices at the layer 1 of some firing graph, with the same input domain $\mathcal{G}$ such that $I = J \cup K$ and $J \cap K = \emptyset$. result (\ref{prop:partition-1}) stands that $\forall x \in F_2(S(\mathcal{G}), \mathcal{G})$

\begin{equation*}
\mathcal{P}_I^{l_0}\left[ x \right] = \sum_{l=l_0}^{\vert I \vert} \sum_{j=0}^{\vert J \vert} \mathcal{P}_{J, j}\left[ x \right]  \cdot \mathcal{P}_{K, l - j}\left[ x \right] 
\end{equation*} 

\textbf{Proof.} The statement above can also be written

\begin{equation*}
\mathcal{P}_{I,l} \left[ x \right] = \sum_{j=0}^{\vert J \vert} \mathcal{P}_{J, j}\left[ x \right]  \cdot \mathcal{P}_{K, l - j}\left[ x \right] 
\end{equation*} 

$\forall$ $l\in \{l_0, \ldots, \vert I\vert\}$, now we propose a simple proof by contradiction. Let $l\in \{l_0, \ldots, \vert I\vert\}$, $X \in S(\mathcal{G})$ and $x \in F_2(\{X\}, \mathcal{G})$ such that

\begin{align*}
\mathcal{P}_{I,l} \left[ x \right] &= 1\\
\sum_{j=0}^{\vert J \vert} \mathcal{P}_{J, j}\left[ x \right]  \cdot \mathcal{P}_{K, l - j}\left[ x \right] &= 0
\end{align*}

Yet, if $J$ and $K$ is a partition of $I$ and $\vert I \cap X \vert = l$, then $(j^{*}, k^{*}) \in \{0, \ldots \vert J \vert \} \times \{ 0, \ldots \vert K \vert\}$ exists such taht

\begin{align*}
\vert X \cap J \vert &= j^{*}\\
\vert X \cap K \vert &= k^{*}\\
j^{*} + k^{*} = l
\end{align*}

Thus for $x \in  F_2(\{ X \}, \mathcal{G})$ 

\begin{align*}
\mathcal{P}_{J, j^{*}}\left[ x \right] \cdot \mathcal{P}_{K, k^{*}}\left[ x \right] = 1\\
\end{align*}

which contradicts our first assumption. Let $l \in \{l_0, \ldots, \vert I \vert\}$, $X \in S(\mathcal{G})$ and $x \in F_2(\{X\}, \mathcal{G})$ such that

\begin{align*}
\mathcal{P}_{I,l} \left[ x \right] &= 0\\
\sum_{j=0}^{\vert J \vert} \mathcal{P}_{J, j}\left[ x \right]  \cdot \mathcal{P}_{K, l - j}\left[ x \right] &= 1
\end{align*}

Thus above statement implies that $j^{*} \in \{0, \ldots, min(l, \vert J \vert)\}$ exists such that  

\begin{equation*}
\mathcal{P}_{J, j^{*}}\left[ x \right]  \cdot \mathcal{P}_{K, l - j^{*}}\left[ x \right] = 1
\end{equation*}

Thus $\vert X \cap J \vert = j^{*}$ and $\vert X \cap K \vert = l - j^{*}$. Since $J$ and $K$ is a partion of $I$ we must have

\begin{equation*}
\vert X \cap I \vert = \vert X \cap J \vert + \vert X \cap K \vert = l
\end{equation*}

As a consequence for $x \in F_2(\{X\}, \mathcal{G})$, $\mathcal{P}_{I,l} \left[ x \right] = 1$ and give us the contradiction. 
\begin{center}
\rule[0pt]{100pt}{1pt} 
\end{center}

Result (\ref{prop:partition-2}) is a particular case of result (\ref{prop:partition-1})\\

\underline{Decomposition}\\

Let $G$ be a firing graph with layer 0 composed of $\mathcal{G}$. Let $u(I, l_u)$, $v(I', l_v)$ such that $I \cap I' = \emptyset$ be vertices of layer 1 and $w(\{ u, v \}, 2)$ be a vertex of layer 2. Let $K \in \cup_{l \in \{l_v, \ldots, \vert I' \vert \}} S(I', l)$, $x \in F_2(S(\mathcal{G}), \mathcal{G})$ and $x' = \begin{bmatrix} \mathcal{P}_{u}[x] &\mathcal{P}_{v}[x] \end{bmatrix}$, the result (\ref{prop:2layer-1}) stands that

\begin{equation*}
\mathcal{P}_{K, \vert K \vert}\left[ x \right] \cdot \mathcal{P}_{\{u, v\}, 2}\left[ x' \right] = \sum_{l=l_u}^{\vert I \vert}  \sum_{J \in S(I, l)} \mathcal{P}_{J \cup K, l + \vert K \vert }\left[ x \right]
\end{equation*}

\textbf{Proof.} The proof the above statement is derived by a straight forward development of the equation, first using result (\ref{prop:2layer-1}) and the fact that $K$ and $I' \setminus K$ is a partition of $I'$ we can write 

\begin{align*}
\mathcal{P}_{\{u, v\}, 2} \left[ x' \right] &= \left( \sum_{l=l_u}^{\vert I \vert} \mathcal{P}_{I, l} \left[ x \right] \right) \cdot \left( \sum_{l=l_v}^{\vert I' \vert} \mathcal{P}_{I', l}\left[ x \right] \right) \\
&= \left( \sum_{l=l_u}^{\vert I \vert} \mathcal{P}_{I, l}\left[ x \right] \right) \cdot \left( \sum_{l=l_v}^{\vert I' \vert}  \sum_{k=0}^{\vert K \vert} \mathcal{P}_{K, k}\left[ x \right] \cdot \mathcal{P}_{I' \setminus K , l-k}\left[ x \right] \right) \\
\end{align*}

Thus 

\begin{align*}
\mathcal{P}_{K, \vert K \vert}\left[ x \right] \cdot \mathcal{P}_{\{u, v\}, 2} \left[ x' \right] &= \mathcal{P}_{K, \vert K \vert} \left[ x \right] \cdot \left( \sum_{l=l_u}^{\vert I \vert} \mathcal{P}_{I, l} \left[ x \right] \right) \cdot \left( \sum_{l=l_v}^{\vert I' \vert}  \sum_{k=0}^{\vert K \vert} \mathcal{P}_{K, k} \left[ x \right] \cdot \mathcal{P}_{I' \setminus K , l-k} \left[ x \right] \right) \\
&= \left( \sum_{l=l_u}^{\vert I \vert} \mathcal{P}_{I, l}\left[ x \right] \right) \cdot \left( \sum_{l=l_v}^{\vert I' \vert}  \mathcal{P}_{K, \vert K \vert}\left[ x \right] \cdot \mathcal{P}_{I' \setminus K , l- \vert K \vert} \left[ x \right] \right)\\
&= \left( \sum_{l=l_u}^{\vert I \vert} \mathcal{P}_{I, l}\left[ x \right] \cdot \mathcal{P}_{K, \vert K \vert}\left[ x \right] \right) \cdot \underbrace{ \sum_{l=0}^{\vert I' \vert - \vert K \vert} \mathcal{P}_{I' \setminus K , l}\left[ x \right]}_{=1}\\
&= \sum_{l=l_u}^{\vert I \vert} \mathcal{P}_{I, l}\left[ x \right] \cdot \mathcal{P}_{K, \vert K \vert}\left[ x \right]
\end{align*}

The last line is equal to $\sum_{l=l_u}^{\vert I \vert}  \sum_{J \in S(I, l)} \mathcal{P}_{J \cup K, l + \vert K \vert }\left[ x \right]$ and thus the proof is achieved.

\begin{center}
\rule[0pt]{100pt}{1pt} 
\end{center}

Result (\ref{prop:2layer-1}) is a particular case of result (\ref{prop:2layer-2})\\

Let $G$ be a firing graph with layer 0 composed by measure grid's bits $\mathcal{G}$ and $f \in \mathcal{F}$ denote some target factor that is linked to some bit of the measure grid. The distribution of activation of latent factors and measure grid's bits will be denoted  $d$ and $d'$ and the event "factor $f$ is active" will be denoted by $e$. Furthermore, let $v$ be some vertex of $G$ whose characteristic polynome respects $\mathcal{P}_v = \mathcal{P}_{I}^{l}$ with $(I, l) \in S(\mathcal{G}), \{1, \ldots, \vert I \vert\}$ and $f \in \mathcal{F}$ some factor.\\

\underline{Precision of vertex}\\

The result (\ref{prop:precision1}) stands that the precision of $v$ with respect to $f$ writes

\begin{equation*}
\phi_{v, f} = \frac{\Vert \mathcal{P}_{f} \Vert_{d'}}{\Vert \mathcal{P}_{f} \Vert_{d'} + (1 - \Vert \mathcal{P}_{f} \Vert_{d'}) \times \omega_{I, l, f}}
\end{equation*}

\textbf{Proof.} First, starting from the defintion of $\phi_{v, f}$

\begin{equation*}
\phi_{v, f} = \phi_{I, l, f} = \frac{\Vert \mathcal{P}_{I}^{l} \Vert_{d, e}}{\Vert \mathcal{P}_{I}^{l} \Vert_{d}}
\end{equation*}

Thus using $\mathcal{P}_{*} = \mathcal{P}_{\mathcal{G}(f), \vert \mathcal{G}(f) \vert}$ one have

\begin{equation*}
\phi_{v, f} = \frac{\langle \mathcal{P}_{I}^{l}, \mathcal{P}_{*} \rangle_{d, e} + \langle \mathcal{P}_{I}^{l}, \bar{\mathcal{P}}_{*} \rangle_{d, e} }{\langle \mathcal{P}_{I}^{l}, \mathcal{P}_{*} \rangle_{d, e} + \langle \mathcal{P}_{I}^{l}, \bar{\mathcal{P}}_{*} \rangle_{d, e} + \langle \mathcal{P}_{I}^{l}, \mathcal{P}_{*} \rangle_{d, \bar{e}} + \langle \mathcal{P}_{I}^{l}, \bar{\mathcal{P}}_{*} \rangle_{d, \bar{e}}}
\end{equation*}

Yet $\langle \mathcal{P}_{I}^{l}, \bar{\mathcal{P}}_{*} \rangle_{d, e} = 0$

\begin{equation*}
\phi_{v, f} = \frac{\langle \mathcal{P}_{I}^{l}, \mathcal{P}_{*} \rangle_{d \vert e} \times \Vert \mathcal{P}_{f} \Vert_{d'} }{\langle \mathcal{P}_{I}^{l}, \mathcal{P}_{*} \rangle_{d \vert e} \times \Vert \mathcal{P}_{f} \Vert_{d'} + \left( \langle \mathcal{P}_{I}^{l}, \mathcal{P}_{*} \rangle_{d \vert \bar{e}} + \langle \mathcal{P}_{I}^{l}, \bar{\mathcal{P}}_{*} \rangle_{d \vert \bar{e}} \right)\times \Vert \bar{\mathcal{P}}_{f} \Vert_{d'}} 
\end{equation*}

Finally by identification of term

\begin{equation*}
\phi_{v, f} = \frac{\mu_{I, l, f} \times \Vert \mathcal{P}_{f} \Vert_{d'} }{\mu_{I, l, f} \times \Vert \mathcal{P}_{f} \Vert_{d'} + \nu_{I, l, f} \times \left( 1 -  \Vert \mathcal{P}_{f} \Vert_{d'} \right)}
\end{equation*}

Which gives the expected result.

\begin{center}
\rule[0pt]{100pt}{1pt} 
\end{center}

The result (\ref{prop:precision2}) stands that if $\mu_{v, f} = 1$ we have

\begin{equation*}
\phi_{v, f} \leq \frac{\Vert \mathcal{P}_{f} \Vert_{d'}}{\Vert \mathcal{P}_{f} \Vert_{d'} + (1 - \Vert \mathcal{P}_{f} \Vert_{d'}) \times \omega_{\mathcal{G}(f),\vert \mathcal{G}(f) \vert, f}} 
\end{equation*}

\textbf{Proof.} The result (\ref{prop:precision2}) can be proven by simple contradiction, suppose there is a tuple $(I, l_0) \neq (\mathcal{G}(f), \vert \mathcal{G}(f) \vert)$ such that 

\begin{align*}
\mu_{I, l_0, f} &= 1 \\
\nu_{I, l_0, f} &< \nu_{\mathcal{G}(f), \vert \mathcal{G}(f) \vert, f}
\end{align*}

First, denoting $\mathcal{P}_{\mathcal{G}(f),\vert \mathcal{G}(f) \vert} = \mathcal{P}_{*}$ and using (\ref{prop:partition-1}) we have

\begin{align*}
\langle  \mathcal{P}_{I}^{l_0}, \mathcal{P}_{*}\rangle_{d} &=\sum_{l=l_0}^{\vert I \vert} \sum_{k=0}^{I\cap \mathcal{G}(f)} \langle \mathcal{P}_{I\cap \mathcal{G}(f), k} \cdot  \mathcal{P}_{I\setminus \mathcal{G}(f), l-k}, \mathcal{P}_{*} \rangle_{d} \\
&= \sum_{l=l_0}^{\vert I \vert} \sum_{k=0}^{I\cap \mathcal{G}(f)} \sum_{x \in F_2(S(\mathcal{G}), \mathcal{G})} \mathcal{P}_{I\cap \mathcal{G}(f), k}[x] \cdot  \mathcal{P}_{I\setminus \mathcal{G}(f), l-k} [x] \cdot \mathcal{P}_{*}[x] \times d_{x}\\
&= \sum_{x \in F_2(S(\mathcal{G}), \mathcal{G})}  \mathcal{P}_{*}[x] \cdot \left( \sum_{l=l_0}^{\vert I \vert} \mathcal{P}_{I\setminus \mathcal{G}(f), l-\vert I \cap \mathcal{G}(f)\vert} [x] \right) \times d_{x}
\end{align*}

Where $d$ is any well defined ditribution on $F_2(S(\mathcal{G}), \mathcal{G})$. Thus, we have 

\begin{equation*}
\mu_{I, l_0, f} = \sum_{x \in F_2(S(\mathcal{G}), \mathcal{G})}  \mathcal{P}_{*}[x] \cdot \left( \sum_{l=l_0}^{\vert I \vert} \mathcal{P}_{I\setminus \mathcal{G}(f), l-\vert I \cap \mathcal{G}(f)\vert} [x] \right) \times d_{x \vert e}
\end{equation*}

As a consequence, in order to have $\mu_{I, l_0, f} = 1$ we most have $ \lbrace \mathcal{P}_{I\setminus \mathcal{G}(f), l-\vert I \cap \mathcal{G}(f)\vert} \rbrace_{l \in \{ l_0, \ldots, \vert I \vert \}}$ to be a partion of $F_2(S(\mathcal{G}), \mathcal{G})$. Thus $l_0 \leq \vert I \setminus \mathcal{G}(f) \vert$. On the other end the precision coefficient writes

\begin{equation*}
\nu_{I, l_0, f} = \sum_{x \in F_2(S(\mathcal{G}), \mathcal{G})}  \mathcal{P}_{*}[x] \cdot \left( \sum_{l=l_0}^{\vert I \vert} \mathcal{P}_{I\setminus \mathcal{G}(f), l-\vert I \cap \mathcal{G}(f)\vert} [x] \right) \times d_{x \vert \bar{e}} +  \langle  \mathcal{P}_{I}^{l_0}, \bar{\mathcal{P}}_{*}\rangle_{d \vert \bar{e}}
\end{equation*}

so if $l_0 \leq \vert I \setminus \mathcal{G}(f) \vert$

\begin{align*}
\nu_{I, l_0, f} &=  \nu_{\mathcal{G}(f), \vert \mathcal{G}(f) \vert, f} + \langle  \mathcal{P}_{I}^{l_0}, \bar{\mathcal{P}}_{*}\rangle_{d \vert \bar{e}}\\
&\geq \nu_{\mathcal{G}(f), \vert \mathcal{G}(f) \vert, f}
\end{align*}

Since $\langle  \mathcal{P}_{I}^{l_0}, \bar{\mathcal{P}}_{*}\rangle_{d \vert \bar{e}} \geq 0$, which lead to a contradiction.

\begin{center}
\rule[0pt]{100pt}{1pt} 
\end{center}

\underline{Recall of vertex}\\

The result (\ref{prop:recall1}) stands that the recall of $v$ with respect to $f$ is

\begin{equation*}
\psi_{v, f} = \mu_{I, l, f}
\end{equation*}

Furthermore, the result (\ref{prop:recall2}) stands that 

\begin{equation*}
0 \leq \phi_{v, f} \leq 1
\end{equation*}

Where right equality is reached whenever $v$ is connected to a set of measure grid's bit $I \in \mathcal{G}$, with level $l_0 = \vert I \vert$ such that $I \subset \mathcal{G}(f)$. \\ 

\textbf{Proof.} From the definition of $\psi_{v, f}$ we have \\

\begin{equation*}
\psi_{v f} = \psi_{I, l, f} = \frac{\Vert \mathcal{P}^{l}_I \Vert_{d, e}}{\Vert \mathcal{P}_{\mathcal{G}(f)} \Vert_{d, e}}
\end{equation*}

Thus using $\mathcal{P}_{*} = \mathcal{P}_{\mathcal{G}(f), \vert \mathcal{G}(f) \vert}$ one have

\begin{equation*}
\psi_{v f} = \frac{\langle \mathcal{P}^{l}_I, \mathcal{P}_{*} \rangle_{d, e} + \langle \mathcal{P}^{l}_I, \bar{\mathcal{P}}_{*} \rangle_{d, e}}{\Vert \mathcal{P}_{f} \Vert_{d'}}
\end{equation*}

Yet $\langle \mathcal{P}^{l}_I, \bar{\mathcal{P}}_{*} \rangle_{d, e} = 0$, thus 
 
\begin{align*}
\psi_{v f} = \frac{\langle \mathcal{P}^{l}_I, \mathcal{P}_{*} \rangle_{d, e}}{\Vert \mathcal{P}_{f} \Vert_{d'}} = \frac{\langle \mathcal{P}^{l}_I, \mathcal{P}_{*} \rangle_{d \vert e} \times \Vert \mathcal{P}_f \Vert_{d'}}{\Vert \mathcal{P}_{f} \Vert_{d'}} = \mu_{I, l, f} 
\end{align*}

Finally the result (\ref{prop:recall2}) directly comes with the definition.
\begin{center}
\rule[0pt]{100pt}{1pt} 
\end{center}

\underline{vertex's score process}\\

Let $s_{v, f}[N, T, p, q]$ be the score process of $v$ with respect to $f$, for some  $N, T, p, q \in \mathbb{N}^4$. The result (\ref{prop:score_mean}) stands that

\begin{equation*}
\mathbb{E} \left[ s_{v, f}[N, T, p, q]  \right] = N + T \times (q_s\times (p + q) - p) = N + T \times (\phi_{I, l, f} \times (p + q) - p)
\end{equation*}

\textbf{Proof.} From the definition of the score process we have 

\begin{equation*}
s_{v, f}[N, T, p, q] = N + \sum_{t=1}^{T} s_{v, p, q, f, t}
\end{equation*}

with $\lbrace s_{v, p, q, f, t} \rbrace_{t=1, \ldots, T}$ a sequence of i.i.d such that

\begin{center}
$s_{v, p, q, t, f} = \begin{cases} q, & \text{with probability } q_s = \frac{q_r}{q_r + q_p}  \\ -p, & \text{with probability } 1 - q_s \end{cases}$
\end{center}

Thus we can write

\begin{align*}
\mathbb{E}\left[ s_{v, f}[N, T, p, q \right] &= N + \sum_{t=1}^{T} \mathbb{E} \left[ s_{v, p, q, t, f}\right]\\
&= N + \sum_{t=1}^{T} q \times q_s - p \times (1 - qs)\\
&= N + \sum_{t=1}^{T} q_s \times (p + q) - p \\
&= N + T \times \left(q_s \times (p + q) - p \right) 
\end{align*}

Yet $q_s = \frac{q_r}{q_r + q_p}$, with $q_r = \Vert \mathcal{P}_v \Vert_{d, e}$ and $q_p = \Vert \mathcal{P}_v \Vert_{d,\bar{e}}$, thus

\begin{equation*}
q_s = \frac{\Vert \mathcal{P}_v \Vert_{d, e}}{\Vert \mathcal{P}_v \Vert_{d, e} + \Vert \mathcal{P}_v \Vert_{d,\bar{e}}} = \frac{\Vert \mathcal{P}_v \Vert_{d, e}}{\Vert \mathcal{P}_v \Vert_{d}}
\end{equation*}

Which is the definition of $\phi_{v, f}$ that is equal to $\phi_{I,l, f}$ by definition.

\begin{center}
\rule[0pt]{100pt}{1pt} 
\end{center}

Furthermore the result (\ref{prop:score_var}) stands that,

\begin{equation*}
\Var \left[ s_{v, p, q, t, f}  \right] = (p + q)^{2} \times \phi_{I,, l, f} \times (1 - \phi_{I, l, f})
\end{equation*}

\textbf{Proof.} Using the result of previous proof we first compute $\mathbb{E}\left[  s_{v, p, q, t, f}^{2}\right]$

\begin{align*}
\mathbb{E}\left[  s_{v, p, q, t, f}^{2}\right] &= q^{2} \times q_s + p^{2} \times (1 - q_s)\\
&=  q^{2} \times \phi_{v, f} + p^{2} \times (1 - \phi_{v, f})\\
&= \phi_{v, f}\times (q + p)\times (q - p) + p^{2}
\end{align*}

Furthermore we have seen previously that $\mathbb{E}\left[  s_{v, p, q, t, f}\right] = \phi_{v, f}\times (p + q) - p$, thus

\begin{equation*}
\mathbb{E}\left[  s_{v,p, q, t, f}\right]^{2} = \phi_{v, f}\times (p + q) \times \left( \phi_{v,f}\times (p + q) - 2\times p \right) + p^{2}
\end{equation*}

Finally
\begin{equation*}
\Var \left[ s_{v, p, q, t, f}  \right] = \mathbb{E}\left[  s_{v, q, p, t, f}^{2}\right] - \mathbb{E}\left[  s_{v, p, q, t, f}\right]^{2} = (q + p)^{2} \times \phi_{v, f} \times (1 - \phi_{v, f})
\end{equation*}

Which gives the expected result since $\phi_{v, f} = \phi_{I, l, f}$ by definition.
\begin{center}
\rule[0pt]{100pt}{1pt} 
\end{center}

\section{Implementation}

The code that has been used to obtain results of simulations can be found on github at \href{https://github.com/pierreGouedard/deyep}{\textit{https://github.com/pierreGouedard/deyep}} under the branch \textit{publi\_1}. The code is exclusively written in python, is compatible with interpreter python2.7 and python3 and requires python modules numpy and scipy. Code for simulation can be found under

\begin{itemize}
\item tests/signal\_plus\_noise\_1.py
\item tests/signal\_plus\_noise\_2.py
\item tests/signal\_plus\_noise\_3.py
\item tests/sparse\_identification.py
\item tests/sparse\_identification\_2.py
\end{itemize}

where the list below are relative to the root directory of the project. The code in the branch \textit{publi\_1} has not changed since the submision of this paper, however, the code in other branch, notably master, may have been optimised, augmented or refactored.

\newpage
\bibliography{mybib}{}
\bibliographystyle{unsrt}

\end{document}